\documentclass{article}

\usepackage[preprint]{neurips_2025}

\usepackage[utf8]{inputenc} 
\usepackage[T1]{fontenc}    
\usepackage{hyperref}       
\usepackage{url}            
\usepackage{booktabs}       
\usepackage{amsfonts}       
\usepackage{nicefrac}       
\usepackage{microtype}      
\usepackage{xcolor}         
\usepackage{subfiles}       
\usepackage{graphicx}
\usepackage{subfigure}
\usepackage{algorithm}
\usepackage{algorithmic}
\usepackage{amsmath}
\usepackage{amssymb}
\usepackage{mathtools}
\usepackage{amsthm}
\usepackage{enumitem}
\usepackage{comment}
\usepackage{afterpage,natbib,lipsum}
\usepackage{wrapfig}
\usepackage{eqparbox}

\usepackage{listings}
\usepackage[T1]{fontenc}
\usepackage{tcolorbox}
\tcbuselibrary{skins,breakable}
\definecolor{LGMaroon}{RGB}{200,0,0}
\definecolor{LGGray}{RGB}{200,200,200}
\newtcolorbox{analysisbox}[1][]{
    enhanced jigsaw,
    colback=white,
    colframe=LGMaroon!75!black,
    fonttitle=\bfseries,
    boxsep=5pt,
    left=5pt,
    right=5pt,
    top=5pt,
    bottom=5pt,
    title=#1
}

\usepackage{multirow}
\title{SMaRT: Select, Mix, and ReinvenT — A Strategy Fusion Framework for LLM-Driven Reasoning and Planning}

\author{%
  Nikhil Verma \\
  \texttt{nikhil.verma@lge.com} \\
  \And
  Manasa Bharadwaj \\
  \texttt{manasa.bharadwaj@lge.com} \\
  \And
  Wonjun Jang \\
  \texttt{strutive07@gmail.com} \\
  \And
  Harmanpreet Singh \\
  \texttt{harmanpreet.singh@lge.com} \\
  \And
  Yixiao Wang \\
  \texttt{Yixiao.wang@lge.com} \\
  \And
  Homa Fashandi \\
  \texttt{homa.fashandi@lge.com} \\
  \And
  Chul Lee \\
  \texttt{clee.lee@lge.com} \\
}

\begin{document}

\maketitle

\begin{abstract}
Large Language Models (LLMs) have redefined complex task automation with exceptional generalization capabilities. 
Despite these advancements, state-of-the-art methods rely on single-strategy prompting, missing the synergy of diverse reasoning approaches. 
No single strategy excels universally, highlighting the need for frameworks that fuse strategies to maximize performance and ensure robustness.
We introduce the \emph{Select, Mix, and ReinvenT}~(SMaRT), an innovative strategy fusion framework designed to overcome this constraint by creating balanced and efficient solutions through the seamless integration of diverse reasoning strategies.
Unlike existing methods, which employ LLMs merely as evaluators, SMaRT uses them as intelligent integrators, unlocking the “\textit{best of all worlds}” across tasks.
Extensive empirical evaluations across benchmarks in reasoning, planning, and sequential decision-making highlight the robustness and adaptability of SMaRT.
The framework consistently outperforms state-of-the-art baselines in solution quality, constraint adherence, and performance metrics.
%
%
This work redefines LLM-driven decision-making by pioneering a new paradigm in cross-strategy calibration, unlocking superior outcomes for reasoning systems and advancing the boundaries of self-refining methodologies.
\end{abstract}
\section{Introduction}\label{intro}


Optimizing LLM inputs and improving their responses is a major research area aimed at improving reasoning capabilities beyond merely scaling up models~\cite{yin2024enhancing}. 
Input optimization strategies, i.e., prompting techniques, have been developed to utilize tailored prompts to guide models in addressing complex tasks, including multi-step and context-sensitive problems. 
These strategies mimic human-like decision-making, enabling the adaption of LLMs to tasks that demand meticulous reasoning and planning for precise completion. 
Prompting strategies have evolved from simple directive-based approaches like few-shot learning \cite{brown2020language} to more sophisticated methodologies such as multi-step reasoning \cite{wei2022chain, yang2024largelanguagemodelsoptimizers}, abstract thinking \cite{zheng2023take}, subproblem decomposition \cite{wang2023plan}, top-down planning \cite{zhou2022least} and self-improvement techniques~\cite{shinn2023reflexion, madaan2023self, chen2024self}.
%
%
%

\begin{wrapfigure}{c}{0.40\linewidth}
   \hfill
 \includegraphics[scale=0.20]{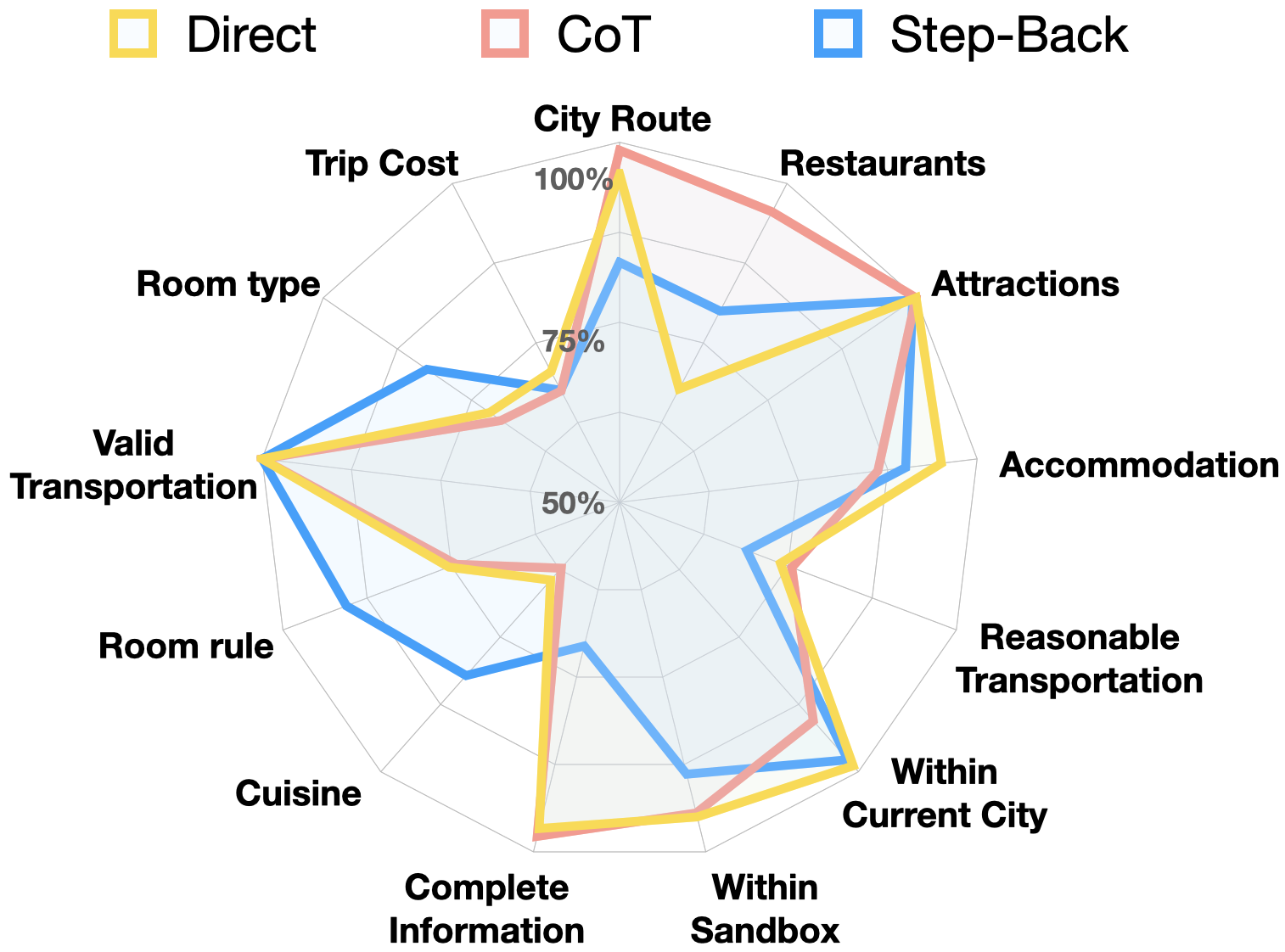}
  \caption {Complex real-world problems often involve multiple constraints that must be satisfied to generate a valid solution. For example, TravelPlanner~\cite{xie2024travelplanner} can involve up to thirteen constraints, from choosing a reasonable city route to a budget-friendly trip. As illustrated in the radar plot, no single prompting technique
  can effectively optimize all these constraints simultaneously.  
  }
    \label{fig:travelplanner_constraint_following}
\end{wrapfigure}

Each strategy presents distinct strengths and limitations, complicating the selection of the most suitable approach for a given task. 
In complex reasoning domains—such as mathematical problem solving—intermediate steps often act as constraints that must be satisfied to reach a correct solution. 
This challenge is amplified in multi-step, long-horizon planning tasks, where multiple constraints must be jointly addressed. For instance, this is evident in TravelPlanning \cite{xie2024travelplanner}, which serves as a representative example of a real-world planning scenario, where success means satisfying multiple requirements.
Using the same language model (Gemini-1.5 \cite{team2024gemini}) with three different strategies—Direct, CoT, and Step Back—the success of each strategy was evaluated across various dimensions of constraint adherence during plan formation, as illustrated in Figure~\ref{fig:travelplanner_constraint_following}.
The observations highlight that each strategy excels in certain dimensions while underperforming in others. 
This creates a challenge, as an effective plan must achieve balanced performance across all critical dimensions.

%
Existing methods for automatic strategy selection either rely on LLM-as-a-judge techniques to evaluate and rank outputs across tasks~\cite{zheng2023judging, xie2023decomposition, li2024generation}, or repeatedly sample a single reasoning strategy and aggregate the results~\cite{wangself2023, zhao2023automatic}.
However, these methods often fall short in leveraging the complementary strengths of multiple strategies to construct a unified, superior solution. Building on the concepts of multi-sampling at inference time and the LLM-as-a-Judge framework, this work introduces the strategy fusion framework called Select, Mix, and ReinvenT (SMaRT).
%
%
%
%
In this framework, instead of merely selecting from pre-generated solutions, the LLM is given the opportunity to \textit{rethink} initial diverse reasoning traces and synthesize a final, curated response by analyzing and integrating the most promising elements across strategies.

\begin{figure*}
\centering
  \includegraphics[scale=0.120]{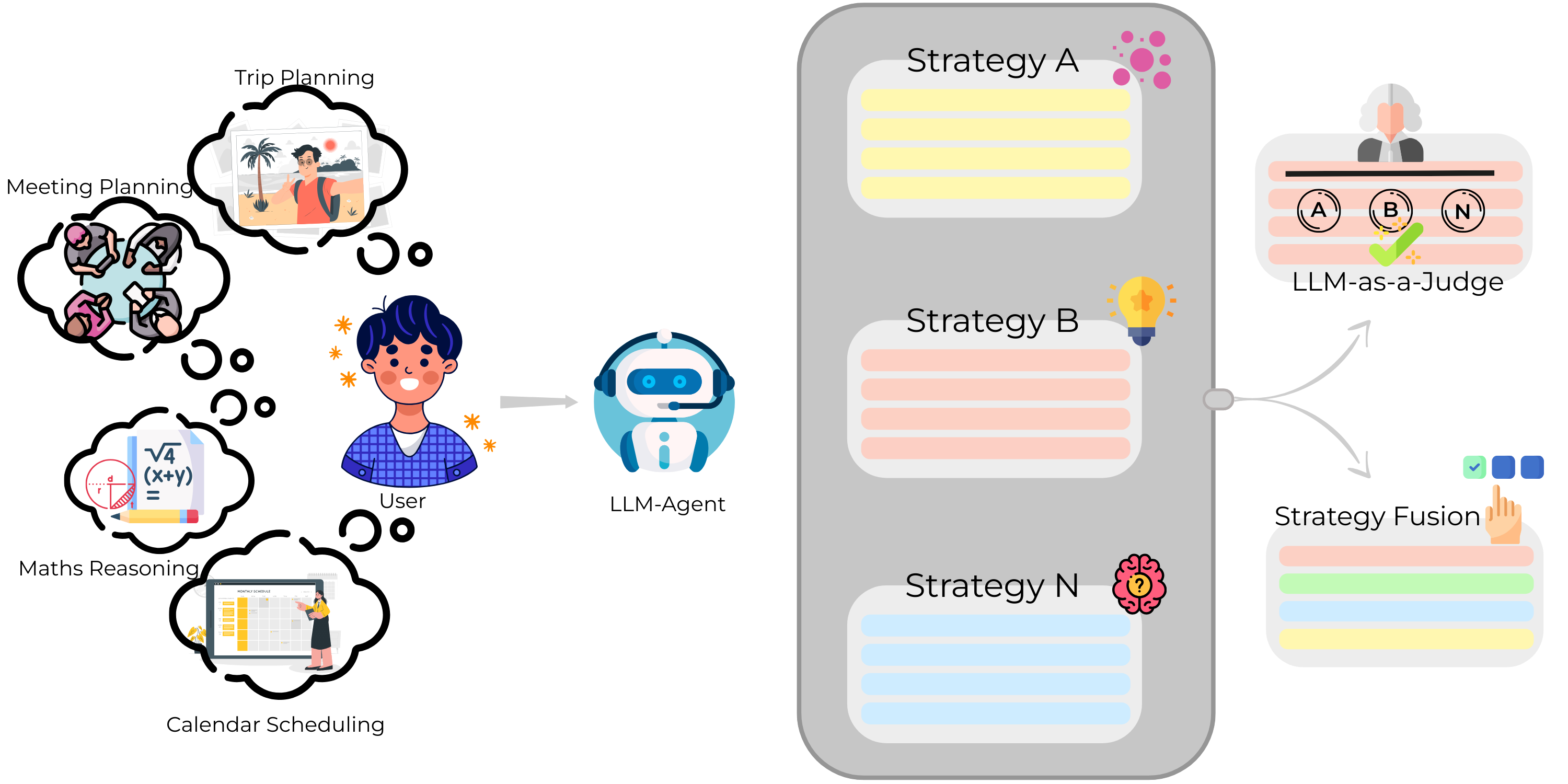}
    \caption {The proposed framework \emph{Select, Mix, and ReinvenT}~(SMaRT) for strategy fusion. To solve a task, the language model first employs base strategies to generate initial solutions based on the input information. In the second stage, the model has the chance to rethink the initial approaches and to develop a new solution. This new solution may incorporate elements from various initial solutions or introduce new components compared to the LLM-as-a-Judge or Self-Consistency framework that only chooses one answer.}
    \label{fig:sff_work_flow}
\end{figure*}

The proposed framework, shown in Figure~\ref{fig:sff_work_flow}, was thoroughly evaluated on various planning and reasoning datasets, providing a comprehensive assessment of its effectiveness across different tasks and reasoning strategies. 
%
%
%
%
The results consistently show that our framework's final outputs outperform those derived from individual base strategies and significantly exceed the performance of the baseline approaches, such as LLM-as-a-Judge and self-consistency~\cite{wang2023plan}.
Furthermore, the methodology was extended by employing a hybrid approach, combining the output of smaller open-source LLMs with those of larger API-based LLMs. 

The primary contributions of this paper are as follows:

\begin{itemize}[noitemsep,topsep=3pt,parsep=3pt,partopsep=3pt]
    \item We present a novel framework called \emph{Select, Mix, and ReinvenT}~(SMaRT). This framework allows LLMs to first explore various reasoning strategies, thereby broadening their reasoning capabilities. 
    After this initial exploration, the models can create a final solution by selecting, combining, and reinventing elements from their discoveries.
    
    \item We demonstrate how the intrinsic judgment capabilities of LLMs can effectively fuse and cross-optimize the outputs of various input strategies, surpassing the traditional application of LLMs as judges or marginalizing over repeated sampling.
    
    \item Real-world scenarios such as reasoning, planning, and sequential decision-making highlight the remarkable effectiveness of the SMaRT framework. 
    
\end{itemize}

\section{Strategy Fusion Framework}\label{proposed_sff}


We present a framework for strategy fusion that systematically integrates diverse reasoning strategies to enhance the reasoning abilities and problem-solving skills of LLMs.
The framework operates in two distinct stages to maximize reasoning diversity and solution quality.
In Stage I (Initial Solutions), the model produces candidate solutions using a suite of complementary base strategies.
Stage II (Fusion) evaluates these candidates and synthesizes a final solution by selectively integrating elements from multiple strategies or generating novel components.
This approach enables both rigorous evaluation and creative recombination, surpassing the limitations of single-strategy reasoning.
A formal description of our framework is provided in Algorithm~\ref{alg:sff2}.

\begin{itemize}[noitemsep,topsep=0pt,parsep=0pt,partopsep=0pt]
    \item\textbf{Stage I- Initial Solutions:} The LLM, $\mathcal{L}$, generates $N$ initial solutions for a specific task $T$ by using the task description $d_T$, and a set of $N$ base selection strategies $S$.
    In general, the selected base strategies should be suitable for the specific task, and the task description should provide sufficient information regarding the task. 
    However, there is no restriction on the type of prompting strategy. 
    The proposed framework reduces the reliance on a specific inference-time approach and allows the use of multiple strategies.
    %
    %
    {Stage I requires $N$ inferred outputs from LLM to collect the initial solutions.} 
    
    \item\textbf{Stage II- Fusion:} 
    %
    %
    %
    The initial solutions derived from the base strategies and task description are sent to the LLM using a structured fusion prompt known as the \emph{fuser}, which embeds sufficient context to critically evaluate the initial solutions.
    At this stage, the LLM acts as a reasoning engine, synthesizing a final answer by evaluating and reconciling the diverse outputs produced in Stage I.
    %
    %
    %
    SMaRT assess the components of each solution proposed by diverse strategies, combining them to arrive at the most suitable answer based on its evaluation. 
    If none of the available solutions are adequate, the $fuser$ has the capacity to rethink the problem and create an entirely new solution during the fusion stage.
\end{itemize}

The proposed strategy fusion framework introduces two key generalizations to current techniques. 
First, it reduces the reliance on a single prompting strategy to escape exploration insufficiency. 
The strategy fusion framework can utilize multiple prompting strategies; there is no limit on the number or types of strategies, but it is advisable to keep the selection small to limit the number of inference calls in Stage I.
The second generalization pertains to the LLM-as-a-Judge framework. 
Here, we relax the requirement of selecting just one solution from the available options. 
Instead, we leverage the intrinsic capabilities of the LLM to evaluate the available solutions and arrive at a final solution by mixing them or even creating a new one.

\renewcommand{\algorithmiccomment}[1]{\eqparbox{}{\(\triangleright\) \textit{#1}}}

\makeatletter
\newcommand{\Scale}[2][1]{\scalebox{#1}{$\m@th#2$}}
\makeatother
\newcommand{\sblacktriangleright}{%
  \vcenter{\hbox{\Scale[0.75]{\blacktriangleright}}}%
}
\begin{algorithm}[t]
\caption{Select, Mix, and ReinvenT~(SMaRT)}
\label{alg:sff2}
    \textbf{Input}: 
     \begin{itemize}[noitemsep,topsep=0pt,parsep=0pt,partopsep=0pt]
        \item[$\sblacktriangleright$]   Task $T$ with description $d_{\text{\tiny T}}$
        \item[$\sblacktriangleright$] LLM, $\mathcal{L}$ with a reasoning strategy, $s$,  $\mathcal{L}_{s}$
        \item[$\sblacktriangleright$]  Set of base strategies, 
        $S = \{s_i \mid$ ${i} \in \{2, \ldots, N\},$  where $N \in \mathbb{Z}_{+}$ and $N\geq 2\}$
         \item[$\sblacktriangleright$]  Fusion Strategy, $\text{fuser}$
     \end{itemize}
    \textbf{Output}:
    \begin{itemize}[noitemsep,topsep=0pt,parsep=0pt,partopsep=0pt]
        \item [$\sblacktriangleright$] LLM response, $r$
    \end{itemize}
    
    \begin{algorithmic}[1] 
      
    \STATE Initialize base strategy responses, $R \gets [\ ]$
    \\
    \algorithmiccomment{Stage I: Initial Solutions}
    \FOR{each base strategy $s_i$ in $S$}
        \STATE Generate response $r_i$ for task $T$ based on $d_{\text{\tiny T}}$ using $s_i$
        \STATE $r_{i} \gets \mathcal{L}_{s_i}(d_{\text{\tiny T}})$ 
        \STATE Append $r_i$ to $R$
    \ENDFOR
    \\
    \algorithmiccomment{Stage II: Fusion Stage}
    
        \STATE Generate the final response, $r$, for task $T$ based on $d_T$, by first assessing the initial solutions in $R$, re-think and       arrive at a final solution by combining the available solutions, creating a new component, or selecting one.
        \STATE $r \gets \mathcal{L}_{\text{fuser}}(d_{\text{\tiny T}}, R)$ 
        
    \STATE \textbf{return} $r$
    \end{algorithmic}
\end{algorithm}

\section{Experimental Details and Results}
\label{sec:expt_details}

We conducted experiments on mathematical reasoning, sequential decision-making, and real-world planning scenarios to evaluate the framework{'}s effectiveness across diverse tasks. 
For mathematical reasoning, we utilized the GSM8K~\cite{cobbe2021training} benchmark. 
To evaluate sequential decision-making, the ALFWorld dataset \cite{shridhar2021alfworld} was employed.
Natural Plan \cite{zheng2024natural} and the Travel Planner \cite{xie2024travelplanner} datasets were utilized for planning tasks.
Details of all the datasets are outlined in upcoming sections.

\subsection{Experimental Setup: LLMs and Base Strategies}
\label{subsec:exp_setup}
We rigorously evaluated our framework using two state-of-the-art LLMs trained on diverse datasets and methodologies: the Gemini-1.5 model \cite{team2024gemini} (including its “\textit{flash}" variant) and the GPT-4 model \cite{hurst2024gpt} (including its “\textit{omni}" variant), both representing the latest advancements in their respective releases.
Unless otherwise specified, the same LLM is consistently employed for the base strategy and the strategy fusion processes.
To ensure the reproducibility and consistency of LLM-generated outputs across all experimental settings, the following hyperparameters were meticulously maintained: a deterministic temperature value of 0, a nucleus sampling probability of $top\_p = 0.7$, a token sampling limit of $top\_k = 50$, and a repetition penalty set to 1. 
These parameters were carefully chosen to balance diversity and precision. 
They ensure controlled exploration within the model's probabilistic output space while preserving fidelity to the input context.

We employed multiple prompting strategies, each tailored to the structure and demands of different datasets. 
These included: (1) Direct Prompting, using few-shot examples to guide output format learning; (2) Chain-of-Thought (CoT), which enables step-by-step decomposition of complex tasks; (3) Step-Back Prompting, which abstracts away surface complexity to emphasize high-level reasoning; (4) Plan-to-Code (P2C), a two-stage method that prompts the model to generate structured plans before implementation; and (5) Program-aided Language Models (PAL), which convert natural language into executable code for precise computation. 
For baseline comparisons, we report results using the LLM-as-a-Judge method across all experiments. Additionally, we include comparisons with Self-Consistency, with detailed numbers presented in section~\ref{subsec:ablation}.
Detailed descriptions of prompting strategies are provided in Appendix.
For the reasoning datasets, CoT, PAL, and P2C strategies were employed.
The primary approaches for ALFWorld and Natural Plan (across all three domains) were Direct and CoT strategies.
In the TravelPlanning dataset, Direct, CoT, and Step-Back strategies were utilized. 

\subsection{Mathematical Reasoning}
To evaluate the multi-step reasoning abilities of the proposed approach in mathematical reasoning, we focus on the GSM8K benchmark~\cite{cobbe2021training}. 
GSM8K consists of high-quality, diverse grade-school level math word problems, carefully curated and validated. 
Each problem typically requires multiple reasoning steps, from interpreting the narrative context to deriving intermediate insights and applying appropriate mathematical principles. 
More details on the experimental setup are available in the Appendix~\ref{app:math}. 
%

\textbf{Results:} All of the reasoning task experiments were conducted using a 3-shot setup. 
In the experimental results~(mentioned in Table \ref{tab:gsm8k_results}), we appropriately combined the plain text reasoning from CoT and the plain text outputs from P2C and PAL to derive the final results, and these final results were generated in the form of an executable Python function.

On the GSM8K benchmark with GPT-4, we achieved accuracy scores of 88.5\%, 94.7\%, and 92.2\% using CoT, PAL, and P2C, respectively.
Notably, while the LLM-as-a-Judge approach attained a final accuracy of 88.9\%, the SMaRT framework outperformed all three base strategies, achieving the highest score of 95.5\%.
%
%
For the Gemini-1.5 setting, SMaRT also demonstrates impressive results, achieving a score of 93.6\%. 
It also surpasses CoT, PAL, and P2C, confirming that the synergy among multiple reasoning strategies can benefit smaller-scale or less advanced models. 
The consistent performance improvements highlight the robustness of SMaRT across diverse experimental conditions, underscoring its ability to integrate different lines of reasoning effectively. 
%
%
The results indicate that, in the presence of both effective and ineffective reasoning algorithms, the SMaRT seamlessly integrates multiple approaches, delivering strong performance. 
Despite the significant variance in the performance of individual base strategies, our framework achieves results indistinguishable from the best-performing approach, whereas LLM-as-a-Judge struggles to reach a similar level of approximation.

%
%
%
%

\subsection{Sequential Decision Making}
\label{subsec:expt_alfworld}
ALFWorld is a virtual home navigation environment modeled after the ALFRED embodied agent task dataset~\cite{shridhar2020alfred} and operates as a text-based interactive system.
The embodied tasks are categorized into six types: Pick, Examine, Heat, Cool, Clean, and Pick Two. These tasks involve navigating a home environment to achieve specific goals, such as “\textit{place the vase in the safe}” or “\textit{inspect the book under the desk lamp}.”
Evaluation in ALFWorld is based on the success rate, which measures the number of tasks completed by appropriately organizing and executing the required sub-tasks. More details on the experimental setup are provided in Appendix~\ref{app:alfworld}. Moreover, examples from this dataset can be found in Appendix \ref{app:ex:alfworld_task_samples}.

\begin{table}[t]
  \centering
  \begin{minipage}{0.45\linewidth}
    \centering
    \caption{GSM8K accuracy rates (in \%) using various strategies and LLMs. The best results are shown in bold.}
    \label{tab:gsm8k_results}
    \resizebox{\linewidth}{!}{
      \begin{tabular}{l | c | c}
        \toprule
        \multicolumn{3}{c}{\textbf{GSM8K}} \\ 
        \midrule
        \textbf{Strategy} & \textbf{Gemini-1.5} & \textbf{GPT-4} \\
        \midrule
        \textbf{CoT} & 85.5 & 88.5 \\
        \textbf{PAL} & 92.5 & 94.7 \\
        \textbf{P2C} & 92.9 & 92.2 \\
        \textbf{LLM-as-a-Judge} & 66.2 & 88.9 \\
        \textbf{SMaRT}\textsubscript{\text{ours}} & \textbf{93.6} & \textbf{95.5} \\
        \bottomrule
      \end{tabular}
    }
  \end{minipage}
  \hfill
  \begin{minipage}{0.46\linewidth}
    \centering
    \caption{ALFWorld success rates (in \%) using various strategies and LLMs. The best results are shown in bold.}
    \label{tab:alfworld_results}
    \resizebox{\linewidth}{!}{
      \begin{tabular}{l | c | c}
        \toprule
        \multicolumn{3}{c}{\textbf{ALFWorld Environment}} \\ 
        \midrule
        \textbf{Strategy} & \textbf{Gemini-1.5} & \textbf{GPT-4} \\
        \midrule
        \textbf{Direct} & 55.97 & 76.87 \\
        \textbf{CoT} & 70.89 & 88.81 \\
        \textbf{LLM-as-a-Judge} & 80.59 & 91.79 \\
        \textbf{SMaRT}\textsubscript{\text{ours}} & \textbf{83.58} & \textbf{96.27} \\
        \bottomrule
      \end{tabular}
    }
  \end{minipage}
\end{table}

\textbf{Results:}
The results for all ALFWorld tasks, evaluated using various strategies, are shown in Table \ref{tab:alfworld_results}.
Our SMaRT approach obtains state-of-the-art results by synergizing multiple base responses, achieving task success rates of 83.58\% for Gemini-1.5 and an impressive 96.27\% for GPT-4. 
The results demonstrate that, unlike the base strategies and the LLM-as-a-Judge approach, our SMaRT method combines essential elements from the base plans while effectively identifying and correcting errors in the initial plans.
%


\textbf{Comparison to Reflexion:} Reflexion~\cite{shinn2023reflexion} approach combines natural language feedback with repeated trials to solve failed tasks.
Due to its natural compatibility with sequential decision-making, which leads to incredible gains, we provide an additional comparison of SMaRT to Reflexion.
Error analysis in the Reflexion framework occurs after an attempt (up to 50 turns for ALFWorld~\cite{shinn2023reflexion}) is completed, leading to wasted inferences after an error.
In contrast, our approach seamlessly integrates error analysis and problem-solving into the same attempt.
We conducted fifteen trials for Reflexion with GPT-4, consistent with the original study \cite{shinn2023reflexion}, and obtain a 84.33\% overall success rate. 
Appendix section \ref{app:alfworld_additional_results} presents complete Reflexion results.
In comparison, our method contains a single trial per base method and up to two trials for SMaRT (totaling 3–4 trials).
%
%


\subsection{Natural Plan} 
Natural plan~\cite{zheng2024natural} is the latest benchmark used to assess the effectiveness of LLMs handling planning tasks expressed in natural language. 
The evaluation framework encompasses three distinct and complex tasks: Trip Planning, Meeting Planning, and Calendar Scheduling. Each task is designed to rigorously assess the planning capabilities of language models.

\begin{itemize}[noitemsep,topsep=0pt,parsep=0pt,partopsep=0pt]
    \item \textbf{Trip Planning} involves generating an optimal multi-day itinerary across $N$ European cities, where $N \in [3, 10]$, and determining the duration of stay in each city, with $D \in [2, 7]$. 
    This task tests the model's ability to navigate combinatorial constraints, optimize travel routes, and balance user preferences.
    \item \textbf{Meeting Planning} entails scheduling meetings for $N$ friends, where $N \in [1, 10]$, under constraints such as time availability, location, and mutual compatibility. 
    This scenario requires reasoning to meet multiple people and the resolution of interdependent constraints.
    \item \textbf{Calendar Scheduling} focuses on arranging work meetings for $N$ participants across $D$ workdays, with $N \in [2, 7]$ and $D \in [1, 5]$, while respecting existing schedules and additional constraints. 
    This task challenges the model’s capacity to manage overlapping commitments and dynamic prioritization.
\end{itemize}

Each task is rigorously evaluated using success rate metric that reflect the model{'}s ability to generate feasible and constraint-compliant solutions. 
The design of these benchmarks highlights real-world relevance while offering a granular assessment of task-specific performance. Refer to Appendix \ref{app:natural_plan} for more details.

\textbf{Results:}
Table \ref{tab:natural_plan_results} presents the performance of two base strategies—Direct Prompting and CoT-based step-by-step reasoning—both implemented with 5-shot examples for structured output, as per the original work \cite{zheng2024natural}. 
The results demonstrate that the plans generated by the proposed SMaRT framework consistently outperform the performance of base strategies across all tasks.
Our proposed framework, which integrates and cross-mixes information from multiple base strategies, achieves superior outcomes compared to individual base strategies and LLM-as-a-Judge selection. 
For instance, with Gemini-1.5 (GPT-4), SMaRT achieves success rates of 33.2\% (24.5\%) in trip planning, 27.8\% (49.8\%) in meeting planning, and 58.5\% (72.0\%) in calendar-based scheduling.
Detailed performance improvements across task-specific constraints, such as the number of people, days, or cities, are provided in Appendix \ref{app:natural_plan}, further confirming the effectiveness of SMaRT over alternative approaches.

\begin{table}[h]
    \centering
    \caption{Success rates (in \%) on the Natural Plan benchmark using the base strategies (Direct, CoT), LLM-as-a-Judge, and the proposed SMaRT framework. The best results are shown in bold. Prompts related to these experiments are reported in Appendix~\ref{app:natural_plan}.}
    \label{tab:natural_plan_results}
    \resizebox{\linewidth}{!}{
    \begin{tabular}{l | c c | c c | c c}
        \toprule
        & \multicolumn{2}{c|}{\textbf{Trip Planning}} 
        & \multicolumn{2}{c|}{\textbf{Meeting Planning}} 
        & \multicolumn{2}{c}{\textbf{Calendar Scheduling}} \\
        \midrule
        \textbf{Strategy} & \textbf{Gemini-1.5} & \textbf{GPT-4} 
                         & \textbf{Gemini-1.5} & \textbf{GPT-4} 
                         & \textbf{Gemini-1.5} & \textbf{GPT-4} \\
        \midrule
        \textbf{Direct} & 32.2 & 5.6 & 25.2 & 46.3 & 50.2 & 66.1 \\
        \textbf{CoT} & 31.2 & 9.4 & 26.0 & 46.5 & 57.7 & 70.2 \\
        \textbf{LLM-as-a-Judge} & 29.2 & 6.2 & 26.2 & 45.1 & 57.0 & 69.4 \\
        \textbf{SMaRT}\textsubscript{\text{ours}} & \textbf{33.2} & \textbf{24.5} & \textbf{27.8} & \textbf{49.8} & \textbf{58.5} & \textbf{72.0} \\
        \bottomrule
    \end{tabular}
    }
\end{table}

\subsection{TravelPlanner}

The TravelPlanner benchmark~\cite{xie2024travelplanner} provides a robust and intricate environment for evaluating the long-horizon planning capabilities of LLMs functioning as agents.
This task is designed to naturally incorporate diverse constraints, including common-sense constraints (e.g., selecting diverse restaurants and optimizing city routes) and hard constraints (e.g., adhering to budget limitations and meeting specific user requirements). 
A successful plan must address all these constraints while generating a coherent multi-day, multi-city itinerary for the user.
To assess the quality of generated plans, the evaluation framework includes metrics such as micro and macro constraint pass rates, as well as an overall final pass rate, which measures the proportion of feasible plans that satisfy all constraints. 
Further details are provided in Appendix \ref{app:TravelPlanner}.

\textbf{Results:}
The effectiveness of SMaRT was evaluated on the TravelPlanning task using both test (1000 episodes) and validation (180 episodes) splits of the dataset. 
Tables \ref{tab:travelplanner_test_split_results} and \ref{tab:travelplanner_val_split_results_gemma_9b} present the performance scores of all strategies, with the Final Pass Rate representing the number of plans that satisfy all constraints in their entirety.
SMaRT outperforms the base strategies across all splits of the dataset, demonstrating superior adherence to constraints and overall plan quality using state-of-the-art LLMs.
For final plan creation, our solution leveraged base strategies, including Direct, CoT, and Step-Back reasoning. 
%
%
On the test split with GPT-4, the proposed SMaRT achieved a final pass rate of 37.10\%, representing an absolute improvement of 2.80\% over the best-performing base strategy (34.30\%) and 3.20\% over the LLM-as-a-Judge.
A similar trend is observed on the validation split, where the proposed SMaRT achieves a final performance of 38.33\%, outperforming the best-performing base strategy, Direct (34.44\%), and LLM-as-a-Judge (37.22\%).

\begin{table}[h]
\caption{Performance indicators LLM agent using Direct and CoT on TravelPlanner's test split.}
\centering
\resizebox{\columnwidth}{!}{
\begin{tabular}{l || cccccc || cccccc }
\hline

\multirow{3}{*}{\textbf{Strategy}} &

\multicolumn{1}{c}{\multirow{3}{*}{\begin{tabular}[c]{@{}c@{}}{Delivery}\\{Rate}\end{tabular}}} &
\multicolumn{2}{c}{{Commonsense}} &
\multicolumn{2}{c}{{Hard Constraint}} &
\multicolumn{1}{c}{\multirow{3}{*}{\begin{tabular}[c]{@{}c@{}}{Final}\\{Pass Rate}\end{tabular}}} &

\multicolumn{1}{c}{\multirow{3}{*}{\begin{tabular}[c]{@{}c@{}}{Delivery}\\{Rate}\end{tabular}}} &
\multicolumn{2}{c}{{Commonsense}} &
\multicolumn{2}{c}{{Hard Constraint}} &
\multicolumn{1}{c}{\multirow{3}{*}{\begin{tabular}[c]{@{}c@{}}{Final}\\{Pass Rate}\end{tabular}}}
\\

\multirow{3}{*}{} &
\multicolumn{1}{c}{} &
\multicolumn{2}{c}{{Pass Rate}} &
\multicolumn{2}{c}{{Pass Rate}} &
\multicolumn{1}{c}{} &
\multicolumn{1}{c}{} &
\multicolumn{2}{c}{{Pass Rate}} &
\multicolumn{2}{c}{{Pass Rate}} &
\multicolumn{1}{c}{}
\\

\cmidrule(lr){3-6} \cmidrule(lr){9-12}

\multirow{3}{*}{} &
\multicolumn{1}{c}{} &
{Micro} & {Macro} &
{Micro} & {Macro} &
\multicolumn{1}{c}{} &
\multicolumn{1}{c}{} &
{Micro} & {Macro} &
{Micro} & {Macro} &
\multicolumn{1}{c}{}

\\ \hline \hline

& \multicolumn{4}{c}{with 2 strategies} & \multicolumn{4}{c}{\textbf{Gemini-1.5}} & \multicolumn{4}{c}{with 3 strategies}         \\\hline
\textbf{Direct} &\textbf{100}	& 88.98	& 38.90	& \textbf{70.92}	& \textbf{54.50}	& 23.10 &\textbf{100}	& 88.98	& 38.90	& 70.92	& \textbf{54.50}	& 23.10 \\
\textbf{CoT} &\textbf{100}	& 84.64	& 40.90	& 56.89	& 43.00	& 25.60 &\textbf{100}	& 84.64	& 40.90	& 56.89	& 43.00	& 25.60
\\
\textbf{Step-Back} & - & - & - & - & - & - &\textbf{100}	& 84.34	& 35.60	& 57.12	& 44.80	& 24.60
\\ \hline
\textbf{LLM-as-a-Judge} & \textbf{100} & 86.43 & 44.80 & 62.82 & 45.68 & 26.00 &\textbf{100}	& 88.23	& 45.20	& 66.16	& 47.10	& 26.30 \\ 
\textbf{SMaRT}\textsubscript{\text{ours}} & \textbf{100} & \textbf{89.34} & \textbf{47.40} & 68.76 & 48.50 & \textbf{26.80} &\textbf{100}	& \textbf{90.75}	& \textbf{49.40}	& \textbf{71.05}	& 51.80	& \textbf{27.60} \\ \hline \hline

& \multicolumn{4}{c}{with 2 strategies} & \multicolumn{4}{c}{\textbf{GPT-4}}    & \multicolumn{4}{c}{with 3 strategies}     \\\hline

\textbf{Direct} &\textbf{100}	& 89.88	& 45.90	& 69.65	& 60.40	& 31.90 &\textbf{100}	& 89.88	& 45.90	& 69.65	& 60.40	& 31.90\\
\textbf{CoT} &\textbf{100}	& \textbf{91.96	}& \textbf{52.40}	& \textbf{78.95} & \textbf{63.70}	& 34.30 &\textbf{100}	& 91.96	& 52.40	& \textbf{78.95} & 63.70	& 34.30
\\
\textbf{Step-Back} & - & - & - & - & - & - &\textbf{100}	& 84.08	& 21.30	& 78.73	& \textbf{69.80}	& 16.60
\\ \hline
\textbf{LLM-as-a-Judge} & 100 & 91.22 & 48.42 & 76.90 & 62.68 & 32.86 &\textbf{100}	& 91.72	& 50.70	& 78.47	& 63.60	& 33.90
\\
\textbf{SMaRT}\textsubscript{\text{ours}} & \textbf{100} & 91.70 & 49.60 & 69.52 & 59.60 & \textbf{34.80} &\textbf{100}	& \textbf{93.29}	& \textbf{56.80}	& 78.73	& 63.10	& \textbf{37.10}
\\ \hline \bottomrule
\end{tabular}
}
\label{tab:travelplanner_test_split_results}
\end{table}

\begin{figure}[h]
    \begin{minipage}{.45\textwidth}    
    \centering
    \includegraphics[width=\linewidth]{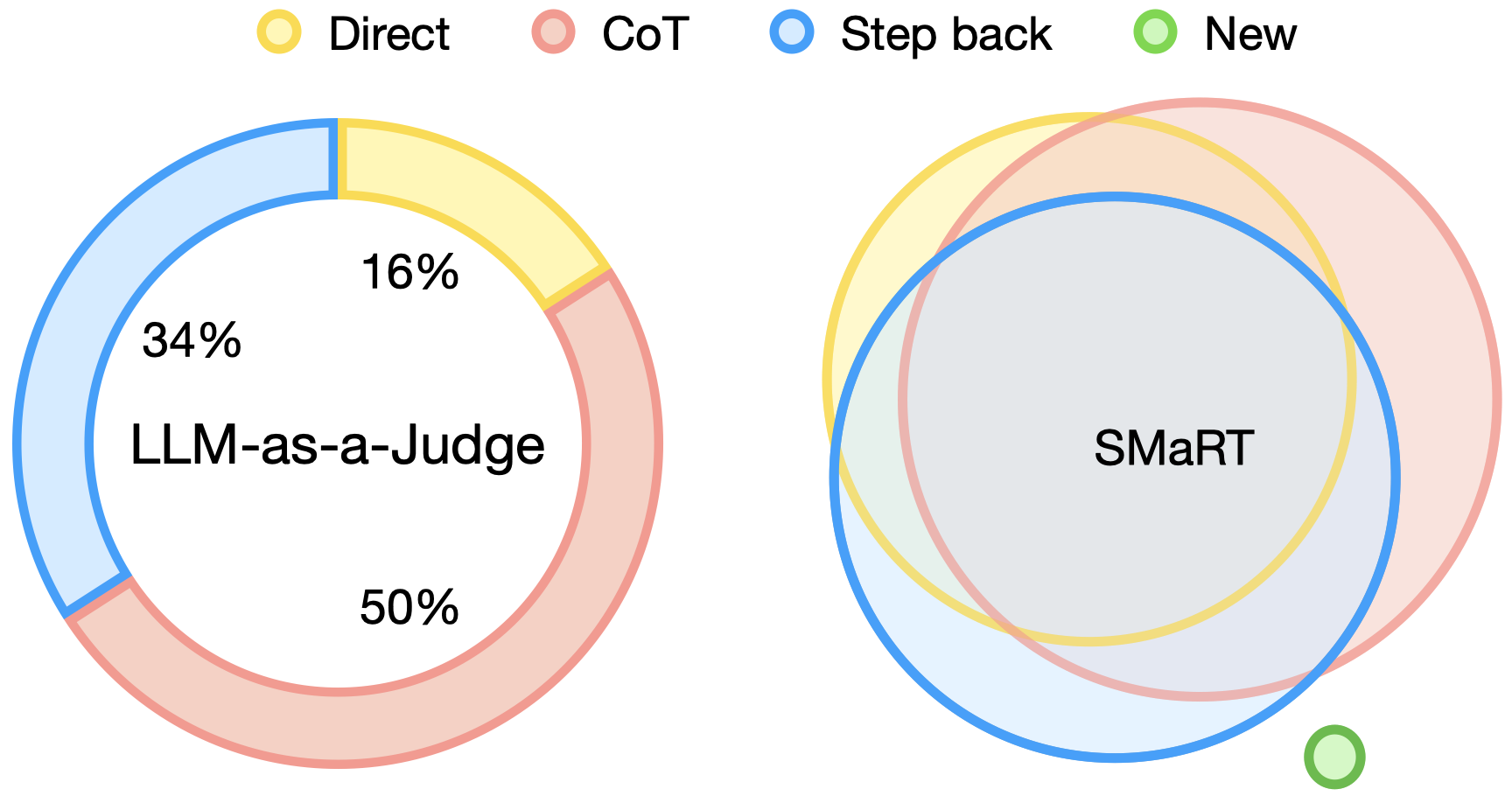}
    \caption{\footnotesize (left) The donut plot displays the percentage of time that the LLM-as-a-Judge selects each strategy. (right) The Venn diagram shows how SMaRT selects components from various base plans for the travel planner benchmark and demonstrates its ability to provide new solutions when existing options fall short. For more details, refer to appendix \ref{app:reasoning}.  }\label{fig:venn}
    \end{minipage}
    \hspace{\fill}
    \begin{minipage}{.50\textwidth}    
    \centering
      \includegraphics[scale=0.2]{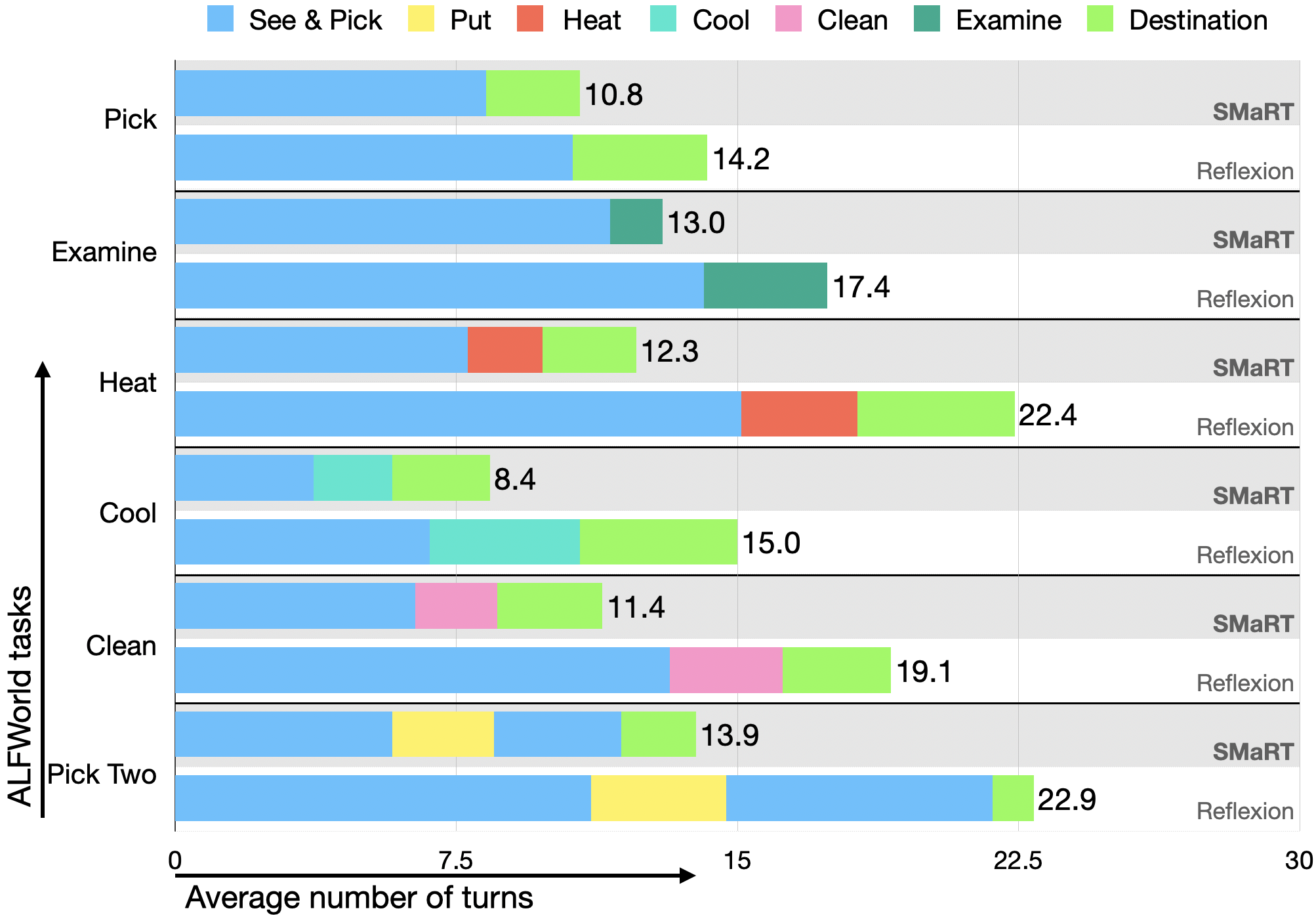}
      \caption {\footnotesize The performance of ALFWorld using GPT-4, highlighting the number of turns taken during the fusion phase of SMaRT and a successful Reflexion trial. The average steps needed to complete both individual sub-tasks and the overall task were analyzed and plotted for the six distinct task categories within this environment. 
    }
        \label{fig:alfworld_sff_reflexion}
    \end{minipage}
\end{figure}

While specific base strategies may excel in certain micro or macro constraint pass rates, SMaRT consistently produces plans that balance and optimize performance across all dimensions.
To visualize this, the performance of LLMs acting as travel agents under various strategies, including the Strategy Fusion Framework, is illustrated in Appendix Figure \ref{fig:tp_base_strat_sff_comparison}, highlighting the capabilities of SMaRT to move towards an optimal solution; 
Refer to Figure~\ref{fig:travelplanner_constraint_following} for comparison with the base strategies. 
The donut diagram in Figure~\ref{fig:venn} (left) displays the percentages of strategies chosen by LLM-as-a-Judge. The Venn diagram in Figure~\ref{fig:venn} (right) shows how strategy fusion frameworks choose different components of each base plan for the travel planner benchmark.
The overlapping region indicates the identical components in multiple base plans. 
The new set shows the SMaRT's rethinking ability to introduce a new solution when none of the initial candidates are deemed satisfactory.
%
%
As observable, most of the SMaRT plan components are selected from different base plans.
Breakfast on the third day is a new addition to the plan, which was not included in any of the base plans but is present in the SMaRT plan. 
%
%
SMaRT provides reasoning for the inclusion or creation of specific components in the final plan; refer to Appendix~\ref{app:reasoning} for the reasoning regarding this example. 

\noindent \textbf{Effect of number of base strategies.} To assess the impact of strategy diversity on the performance of our SMaRT framework, we performed the experiment by varying the number of base strategies used in the TravelPlanner setting. 
Specifically, we evaluate the impact of scaling from two base strategies (Direct and CoT) to three (Direct, CoT, and Step-Back) within our framework.

Our findings, in Table \ref{tab:travelplanner_test_split_results}, indicate that while the two-strategy variant still outperforms the LLM-as-Judge baseline, it consistently underperforms relative to the full three-strategy setup. 
This suggests that increasing the diversity of reasoning strategies provides richer and more complementary perspectives, resulting in improved fused outputs. 
These results highlight the importance of multi-strategy input in enabling more robust final plan selection within SMaRT.

\subsection{Ablation Studies}
\label{subsec:ablation}

\noindent \textbf{Comparison with Self-Consistency.} CoT prompting is a particularly effective base strategy across a range of tasks. 
We further investigated how SMaRT compares to the Self-Consistency framework \cite{wangself2023}, which enhances CoT by aggregating outputs from multiple CoT runs.

We conducted experiments across all datasets and present the comparative results in Table~\ref{tab:self_consistency_comparison}, evaluating three configurations: base CoT, Self-Consistency, and our SMaRT method. 
Our results show that SMaRT consistently outperforms both base CoT and Self-Consistency across most datasets. 
The only exception is Calendar Scheduling, where Self-Consistency slightly edges out SMaRT, though the performance gap remains marginal. 
These results emphasize SMaRT{'}s ability to leverage heterogeneous reasoning styles more effectively than repeated sampling of a single strategy.

\begin{table}[t]
  \centering
  \begin{minipage}{0.40\linewidth}
    \centering
    \caption{Performance comparison of Self-Consistency on the test split across datasets using GPT-4.}
    \label{tab:self_consistency_comparison}
    \resizebox{\linewidth}{!}{
    \begin{tabular}{l || c | c | c}
    \hline
    \textbf{Benchmark} & \textbf{CoT} & \textbf{Self-Consistency} & \textbf{SMaRT}\textsubscript{\text{ours}} \\\hline
    GSM8K & 88.50 & 95.40 & \textbf{95.50} \\
    ALFWorld & 88.81 & 84.32 & \textbf{96.27} \\
    Trip Planning & 9.40 & 6.12 & \textbf{24.50} \\
    Meeting Planning & 46.50 & 42.40 & \textbf{49.80} \\
    Calendar Scheduling & 70.20 & \textbf{73.10} & 72.00 \\
    TravelPlanner & 34.30 & 30.60 & \textbf{37.10} \\\hline
    \end{tabular}}
  \end{minipage}
  \hfill
  \begin{minipage}{0.58\linewidth}
    \centering
    \caption{Performance indicators of LLM agents using various strategies on the TravelPlanner validation split with a combination of models. Best results are in \textbf{bold}.}
    \label{tab:travelplanner_val_split_results_gemma_gemini}
    \resizebox{\linewidth}{!}{
    \begin{tabular}{l || c c c c c c}
    \hline
    \multirow{2}{*}{\textbf{Strategy}} &
    \multirow{2}{*}{\textbf{Delivery Rate}} &
    \multicolumn{2}{c}{\textbf{Commonsense Pass Rate}} &
    \multicolumn{2}{c}{\textbf{Hard Constraint Pass Rate}} &
    \multirow{2}{*}{\textbf{Final Pass Rate}} \\ \cline{3-6}
    & & \textbf{Micro} & \textbf{Macro} & \textbf{Micro} & \textbf{Macro} \\ \hline \hline

    \textbf{Direct}\textsuperscript{Gemma-2-9B} & \textbf{100} & 74.06 & 22.78 & 25.33 & 12.78 & 9.44 \\
    \textbf{CoT}\textsuperscript{Gemma-2-9B} & \textbf{100} & 79.10 & 13.88 & 47.06 & 27.78 & 5.56 \\
    \textbf{Step-Back}\textsuperscript{Gemma-2-9B} & \textbf{100} & 72.50 & 15.56 & 26.32 & 18.89 & 7.78 \\ \hline
    \textbf{LLM-as-a-Judge}\textsuperscript{Gemini-1.5} & \textbf{100} & 78.54 & 23.89 & 42.33 & 23.33 & 10.56 \\
    \textbf{SMaRT}\textsubscript{\text{ours}}\textsuperscript{Gemini-1.5} & \textbf{100} & \textbf{83.61} & \textbf{42.78} & \textbf{49.76} & \textbf{38.33} & \textbf{22.78} \\ \hline
    \bottomrule
    \end{tabular}}
  \end{minipage}
\end{table}

\noindent \textbf{Efficient Strategy Fusion.} To reduce inference costs while preserving performance, we explored hybrid configurations that combine efficient open-source models with high-capacity proprietary models. 
Specifically, we generated base strategy outputs using Gemma-2-9B and delegated the fusion phase to the stronger Gemini-1.5 model. 
This setup leverages the complementary strengths of lightweight and powerful models, showcasing SMaRT{'}s adaptability beyond standard same-model configurations.

Table~\ref{tab:travelplanner_val_split_results_gemma_gemini} presents a detailed comparison. 
While LLM-as-a-Judge with Gemini-1.5 achieved a final pass rate of 10.56\%—on par with a configuration where Gemma-2-9B evaluated its own plans~(Appendix \ref{app:mix_models})—SMaRT substantially outperformed both. 
By synthesizing and enhancing base plans through selective integration and correction, SMaRT reached a 22.78\% pass rate, nearly doubling performance over the best judge-based baseline.

\noindent \textbf{Self-refinement Cost Efficiency.} While inference-time self-refinement strategies like Reflexion prompting concern over multiple LLM calls, our method demonstrates superior efficiency. 
SMaRT achieves higher success rates with significantly fewer refinement trials. 
As shown in Figure~\ref{fig:alfworld_sff_reflexion}, for tasks solved by both methods, SMaRT consistently requires fewer steps and LLM calls across all ALFWorld task types. 
This highlights SMaRT{'}s ability to leverage LLM{'}s planning and self-correction capacities more effectively for scalable sequential decision-making.

Notably, SMaRT was able to generate improved solutions, including novel plan components absent in any individual base plan. 
Our ablation results highlight the promise of strategy fusion in producing high-quality outputs using cost-efficient model combinations, self-refinement capability and bridging the gap between scalability and reasoning performance.

\section{Limitations}
\label{sec:limitations}

SMaRT achieves strong performance through multiple output samples from diverse stratgies, it trades off learning-based adaptability for simplicity. 
Our method does not incorporate internal verification or error-checking mechanisms, and relies on deterministic selection and heuristic fusion rather than learned adaptation. 
Unlike approaches with feedback loops or memory, SMaRT lacks the ability to learn from past mistakes or dynamically adjust over time. 
Additionally, although it avoids retraining and is more efficient than refinement-heavy methods like Reflexion (as shown in section \ref{subsec:ablation}), SMaRT still incurs higher inference costs than simpler baselines such as direct or CoT prompting due to invoking multiple base strategies per task.
%
%
%
%
%
Therefore, a key direction for future work will be to explore optimal selection of base strategies, to integrate memory-based feedback or confidence-weighted fusion mechanisms to enhance adaptability and robustness.

\section{Related Works}\label{litrature}

\textbf{Individual Prompting Strategies.} 
%
Prompting techniques for reasoning and planning span a wide spectrum—from direct instruction and few-shot exemplars to structured reasoning and decomposition. 
Core strategies include Chain-of-Thought (CoT) prompting~\cite{wei2022chain}, Step-Back prompting~\cite{zhengtake2024}, and self-refinement techniques involving reflection and critique~\cite{madaan2023self, dhuliawala2024chain, yao2023react, shinn2023reflexion, jiang2024self}. 
Task decomposition methods such as Least-to-Most~\cite{zhouleast2023}, Plan-and-Solve~\cite{wang2023plan}, Tree-of-Thoughts~\cite{yao2023tree}, and PAL~\cite{gao2023pal} further scaffold complex reasoning.
A comprehensive taxonomy of prompting strategies is provided by \cite{schulhoff2024prompt}.
In our framework, some of these methods serve as modular base strategies. 
While many of them exhibit strong standalone performance, we enhance their effectiveness by incorporating task-specific textual cues (see Appendix sections \ref{app:alfworld}, \ref{app:natural_plan}, \ref{app:TravelPlanner}).

\textbf{Inference-time exploration.}  Inference-time scaling for exploration, augments LLM capabilities by expanding and evaluating multiple reasoning paths at test time \cite{zhang2025and}. 
Early works like self-consistency~\cite{wangself2023} use stochastic sampling with non-zero temperature to generate diverse outputs, but only at the initial step—limiting depth and diversity in reasoning. 
Recent approaches such as Tree-of-Thoughts \cite{yao2023tree}, Graph-of-Thoughts \cite{besta2024graph}, and Forest-of-Thoughts \cite{bi2024forest} extend this idea by branching at intermediate reasoning states, supporting deeper local exploration.
To resolve the resulting multiplicity of outputs, two dominant aggregation strategies have emerged \cite{ke2025survey}: ensemble-based methods, which rely on majority voting or confidence heuristics~\cite{li2024more, wang2024chain}, and verifier-based methods, which use an LLM-as-a-Judge or external evaluators to assess reasoning chains~\cite{setlur2025scaling, qu2025optimizing}. 
Our work differs by unifying exploration and evaluation through strategy fusion, leveraging diverse base prompting strategies (e.g., CoT, Step-Back) and integrating their outputs. 

\textbf{Ensembling.} A wide range of recent work explores inference-time reasoning enhancement through ensembling, self-consistency, and aggregation strategies \cite{schulhoff2024prompt}. 
Self-Consistency\cite{wangself2023} and Universal Self-Consistency\cite{chen2024universal} generate diverse CoT paths by sampling under temperature and aggregate them via majority vote or LLM-based judgment. 
Multi-Chain Reasoning (MCR)\cite{yoran2023answering} advances this idea by prompting the model to meta-reason across multiple CoT chains. 
MoRE\cite{si2023getting} combines outputs from specialized reasoning experts using a Random Forest-based aggregator, but requires task-specific training and feature engineering.
COSP\cite{wan2023better} filters high-agreement CoT chains to form few-shot exemplars, repeating self-consistency for final prediction. 
SuperICL\cite{xu-etal-2024-small} leverages a fine-tuned model to select confident in-context examples, while DENSE~\cite{khalifa2023exploring} averages output probabilities over subsets of demonstrations — both requiring extra tuning or pre-prepared data.

While these approaches rely on either single-strategy sampling, expert-specific modeling, or handcrafted exemplar construction, our proposed framework selects, mixes, and refines diverse base strategies without additional supervision or tuning. 
It offers a modular, self-refining mechanism that consistently yields improved solutions and novel sub-task compositions, demonstrating competitive performance across complex decision-making benchmarks.

\section{Conclusion}
In this work, we propose the framework {Select, Mix, and ReinvenT }~(SMaRT), a novel framework for reasoning and planning that systematically evaluates and integrates components from multiple base strategies to enhance inference-time performance.
Our framework is designed to reinvent an optimal solution which balances between environmental, commonsense, and task-specific constraints. 
Comprehensive evaluations on various benchmarks in mathematical reasoning, planning, and sequential decision-making demonstrate significant improvements in task completion rates and the quality of the generated solutions. 
Notably, SMaRT excels in handling complex, multifaceted tasks that require the seamless coordination of local and global problem-solving strategies. 
Our findings also reveal that leveraging a combination of smaller foundational reasoning modules along with a larger \emph{fuser} model can achieve substantial performance gains with high efficiency.



\bibliographystyle{plain}
\bibliography{neurips_smart_2025_refs}






\clearpage
\clearpage
\appendix

\onecolumn

\section{Reasoning tasks}\label{app:math}
This section provides additional details and experiments on the mathematical reasoning task. First, we list all the prompts in subsection~\ref{app:math:prompts} and include additional experimentation and results on the SVAMP benchmark in subsection~\ref{app:svamp}.

\subsection{Mathematical Reasoning Prompts}\label{app:math:prompts}
This section includes all the prompts used for the mathematical reasoning task. To save space, for prompts with few-shot examples, we have included only one example and omitted the rest in the prompt. Table~\ref{app:math:cot} shows the Chain of Thought (CoT) prompt utilized for the mathematical reasoning task. As mentioned before, the P2C prompting strategy contains two phases, planning phase and coding phase. Table~\ref{app:math:p2c:plan} shows the prompt of the planning phase and Table~\ref{app:math:p2c:code} contains the coding phase prompt. Few shot example of the P2C prompt is shown in Table~\ref{app:math:p2c:shots}. The PAL prompt is shown in Table~\ref{app:math:pal}, this prompt contains the few shot samples as well, only one of them shown. 
Strategy Fusion Framework prompt is shown in Table~\ref{app:math:sff}. 
Tables~\ref{app:math:sff:shots:inputs} and  \ref{app:math:sff:shots:outputs} shows the few shot prompt used in strategy fusion prompt construction. The LLM-as-a-Judge prompt is shown in Table~\ref{app:math:llm:judge}. Table~\ref{app:math:judge:shots} shows the few-shot samples for the LLM-as-a-Judge. 

\begin{table}[h]
\caption{Prompt for CoT on Reasoning tasks}\label{app:math:cot}
\begin{analysisbox}
\textbf{System}:\newline
You are a helpful assistant that can solve math problems step by step.\newline
\textbf{User}:\newline
Let's solve the following math problems. You need to solve these math problems step by step. Here is one example how to do it,\newline
Question: Olivia has \$23. She bought five bagels for \$3 each. How much money does she have left?\newline
\textbf{Assistant}:\newline
Answer:
Olivia had 23 dollars.
And she bought 5 bagels.
And each bagel costs 3 dollars.
So she spent 5 * 3 = 15 dollars.
So she has 23 - 15 = 8 dollars left.
So the answer is 8.\newline
\newline
\{FEW\_SHOTS\}
\newline
\newline
\textbf{User}:\newline
Question: \{QUESTION\}
\end{analysisbox}
\end{table}
\lstset{
    comment=[l]{:},
    commentstyle=\color{blue}, 
    basicstyle=\tiny
}
\lstset{
    backgroundcolor=\color[RGB]{245,245,245},
    breaklines=true,
    breakindent=0pt,
    basicstyle=\ttfamily\small,
    frame=trbl,
    frameround = tttt,
}
\begin{table}[h]
\caption{Prompt for P2C Plan Phase on Reasoning tasks}\label{app:math:p2c:plan}
\begin{analysisbox}
\textbf{System}:\newline
You are writing a plan for a given user instruction with a numbered list. Plan your implementation accordingly.\newline
\textbf{User}:\newline
Write a function to sum the length of the names of a given list of names after removing the names that start with a lowercase letter. Let’s think step by step.\newline
\textbf{Assistant}:\newline
1. Loop the input list.\newline
2. If the name not start with lowercase letter, add the length of the name to result.\newline
3. Return the result.\newline
\textbf{User}:\newline
Write a python function to answer/solve the following: \{QUESTION\} Let's think step by step.
\end{analysisbox}
\end{table}
\begin{table}[h]
\caption{Prompt for P2C Code Phase on Reasoning tasks}\label{app:math:p2c:code}
\begin{analysisbox}
\textbf{User}:
\begin{lstlisting}
def solution():
    '''
    Create a function solution that returns the answer of the following question: {QUESTION}

    Let's think step by step.
    {PLAN}
    '''
\end{lstlisting}
\end{analysisbox}
\end{table}
\begin{table}[h]
\caption{Few shot Prompt for P2C Code phase on Reasoning tasks}\label{app:math:p2c:shots}
\begin{analysisbox}
\textbf{User}:
\begin{lstlisting}
def encrypt(s):
    '''
    Create a function encrypt that takes a string as an argument and returns a string encrypted with the alphabet being rotated. The alphabet should be rotated in a manner such that the letters shift down by two multiplied to two places.
    For example:
    encrypt('hi') returns 'lm' encrypt('asdfghjkl') returns 'ewhjklnop' encrypt('gf') returns 'kj'
    encrypt('et') returns 'ix'

    Let's think step by step.
    1. Create a alphabet, bias two places multiplied by two.
    2. Loop the input, find the latter bias letter in alphabet.
    3. Return result.
    '''
\end{lstlisting}
\textbf{Assistant}:
\begin{lstlisting}
def encrypt(s):
    '''
    Create a function encrypt that takes a string as an argument and returns a string encrypted with the alphabet being rotated. The alphabet should be rotated in a manner such that the letters shift down by two multiplied to two places.
    For example:
    encrypt('hi') returns 'lm' encrypt('asdfghjkl') returns 'ewhjklnop' encrypt('gf') returns 'kj'
    encrypt('et') returns 'ix'

    Let's think step by step.
    1. Create a alphabet, bias two places multiplied by two.
    2. Loop the input, find the latter bias letter in alphabet.
    3. Return result.
    '''
    # Step 1: Create an alphabet, shift it by 2 * 2 = 4 places.
    original_alphabet = 'abcdefghijklmnopqrstuvwxyz'
    shifted_alphabet = original_alphabet[4:] + original_alphabet[:4]

    # Create a dictionary to map each original letter to its encrypted counterpart
    encryption_dict = {original: shifted for original, shifted in zip(original_alphabet, shifted_alphabet)}

    # Step 2: Loop through the input string, find the shifted letter in the alphabet.
    encrypted_string = ''
    for char in s:
        if char in encryption_dict:  # Check if the character is a letter that needs to be encrypted.
            encrypted_string += encryption_dict[char]
        else:
            encrypted_string += char  # If not a letter (e.g., punctuation), leave it as is.

    # Step 3: Return the result.
    return encrypted_string
\end{lstlisting}

\end{analysisbox}
\end{table}
\begin{table}[h]
\caption{Prompt for PAL on Reasoning tasks}\label{app:math:pal}
\begin{analysisbox}
\textbf{System}:\newline
You are a helpful assistant that can solve math problems by using Python code.\newline
\textbf{User}:\newline
Let's use python to solve math problems. You need to write python code to answer these math problems. Here is one example how to do it,\newline
\newline
Problem:\newline
There are 15 trees in the grove. Grove workers will plant trees in the grove today. After they are done, there will be 21 trees.\newline
How many trees did the grove workers plant today?\newline
\newline
\# solution in Python
\newline
\textbf{Assistant}:
\begin{lstlisting}
def solution():
    # Initial number of trees
    initial_trees = 15

    # Final number of trees
    final_trees = 21

    # Calculate the number of trees planted
    trees_planted = final_trees - initial_trees

    return trees_planted
\end{lstlisting}
\{FEW\_SHOTS\}
\newline
\textbf{User}:\newline
\{QUESTION\}
\newline
\newline
\# solution in Python
\end{analysisbox}
\end{table}
\lstset{
    comment=[l]{:},
    commentstyle=\color{blue}, 
    basicstyle=\tiny
}
\lstset{
    backgroundcolor=\color[RGB]{245,245,245},
    breaklines=true,
    breakindent=0pt,
    basicstyle=\ttfamily\small,
    frame=trbl,
    frameround = tttt,
}
\begin{table}[h]
\caption{ Strategy fusion prompt for the mathematical reasoning task}\label{app:math:sff}
\begin{analysisbox}
\textbf{System}:\newline
You are a helpful assistant that can identify the correct answer to the math problem. There are three solutions to the same math problem. I have not yet verified that these solutions are correct. (A) solution uses natural language to answer the question, while the other (B), (C) uses Python program to answer it. (B) directly attempts to answer the problem with coding the solution, while (C) will plan first and then write a code to answer it. You can carefully cross and mix various processes of solving the math problem from each solution to make an optimal solution to the problem, referring to the process you think is correct, and if the provided solution is wrong, modify and improve that part. Before creating your own new solution steps, carefully review each of the steps in (A), (B), and (C), noting which parts are correct and, if any of the steps are incorrect, why they are incorrect, and then create your own new solution steps. Your answer must contains problem solving process and the final answer of problem in python code.
\newline
\newline
\textbf{\# Efficiency Constraints}
\newline
- Time Complexity: The solution should efficiently solve the problem within the smallest possible time complexity, ideally minimizing loops or recursive calls. The expected complexity is either constant O(1), linear O(n), or logarithmic O(logn), depending on the nature of the problem.
- Space Complexity: The solution should use minimal memory, optimizing for O(1) space complexity where feasible. Data structures and variables should not consume excessive space, especially in large input cases.
\newline
\textbf{\# Clarity and Correctness Constraints}
\newline
- Correctness: The solution must accurately and consistently provide the correct answer for all possible valid inputs. Edge cases, such as extreme values or unusual inputs, should be handled appropriately to ensure that the solution is reliable in all scenarios. Comprehensive testing should be conducted to verify the correctness.
- Readability: The solution should be clearly structured and easily understandable. Variable names, comments, and logical sections should be well-defined, ensuring that other programmers can easily follow and maintain the code.
\newline
\textbf{\# Answer Formatting Constraints}
\newline
- Review solutions: This Section should starts with 'Review Solutions: '. Each solution is a process of organizing your thought process before creating the final solution. Please explain in detail which of the steps in each solution are good to reference and which are incorrect thought processes and should be revised and used. Always talk about this review process before creating your final response.
\newline
- Answer Formatting: The final answer should begin with 'Processes: ', followed by an explanation that may include natural language, Python code, or LaTeX syntax. Afterward, 'Final Answer: ' must present only the correct python code solution. Python code solutions must use Markdown Python code blocks (between ```python and ```). This ensures a clear and consistent format for responses. If you generate more then one python function after Final Answer, entrypoint of solution function's name must be 'def solution'.
\newline
- No Natrual Langauge Information After Final Answer: 'Final Answer:' must present only the correct python code solution. DO NOT GENERATE ANY SENTENCES AFTER Final Answer python code.\newline
\textbf{User}:\newline
\{FEW\_SHOTS\}
\newline
Problem: \{QUESTION\}
\newline
(A)
\newline
\{COT\_SOLUTION\}
\newline
(B)
\newline
\{PAL\_SOLUTION\}
\newline
(C)
\newline
\{P2C\_SOLUTION\}

\end{analysisbox}
\end{table}
\begin{table}[h]
\caption{Few shot Prompt for Strategy Fusion on Reasoning tasks (Input)}\label{app:math:sff:shots:inputs}
\begin{analysisbox}
\textbf{User}:\newline
Problem: There are 15 trees in the grove. Grove workers will plant trees in the grove today.After they are done, there will be 21 trees. \newline
How many trees did the grove workers plant today?\newline
\newline
(A)\newline
Answer:\newline
We start with 15 trees. 
Later we have 21 trees.
The difference must be the number of trees they planted.
So, they must have planted 21 - 15 = 6 trees.
Final Answer: The final answer is 6. I hope it is correct.
\newline
\newline
(B)
\begin{lstlisting}
def solution():
    # Initial number of trees
    initial_trees = 15
    
    # Final number of trees
    final_trees = 21
    
    # Calculate the number of trees planted
    trees_planted = final_trees + initial_trees
    
    return trees_planted
\end{lstlisting}
(C)
\begin{lstlisting}
def solution():
    '''
    Create a function solution that returns the answer of the following question: There are 15 trees in the grove. Grove workers will plant trees in the grove today.After they are done, there will be 21 trees.
    How many trees did the grove workers plant today?
    
    Let's think step by step.
    1. Calculate the initial count of the trees.
    2. Subtract the final count from trees.
    3. Return the remaining amount.
    '''
    # Step 1: Calculate the initial count of the trees.
    final_trees = 21
    
    # Step 2: Subtract the final count from trees.
    initial_trees = 15
    remaining_amount = initial_trees - final_trees
    
    # Step 3: Return the remaining amount.
    return remaining_amount
\end{lstlisting}
\end{analysisbox}
\end{table}
\begin{table}[h]
\caption{Few shot Prompt for Strategy Fusion on Reasoning tasks (Output)}\label{app:math:sff:shots:outputs}
\begin{analysisbox}
\textbf{Assistant}:\newline
Review Solutions:\newline
\newline
\textbf{Solution (A):}\newline
This solution correctly identifies that the number of trees planted is the difference between the final and initial number of trees. The calculation \(21 - 15 = 6\) is accurate. The reasoning is sound and leads to the correct answer.\newline
\newline
\textbf{Solution (B):}\newline
In this solution, the code calculates `trees\_planted = final\_trees + initial\_trees`, which sums the initial and final numbers of trees. This is incorrect because it gives \(21 + 15 = 36\), which does not represent the number of trees planted. The correct operation should be subtraction, not addition.\newline
\newline
\textbf{Solution (C):}\newline
This solution attempts to find the number of trees planted by calculating `remaining\_amount = initial\_trees - final\_trees`, resulting in \(15 - 21 = -6\). This yields a negative number, which doesn't make sense in this context. The correct calculation should be `final\_trees - initial\_trees` to get the positive number of trees planted.\newline
\newline
Processes:\newline
\newline
To determine the number of trees planted, subtract the initial number of trees from the final number:\newline
- Initial trees: 15\newline
- Final trees: 21\newline
- Trees planted: `final\_trees - initial\_trees = 21 - 15 = 6`\newline
\newline
This calculation accurately reflects the number of trees planted today.\newline
\newline
Final Answer:
\begin{lstlisting}
def solution():
    initial_trees = 15
    final_trees = 21
    trees_planted = final_trees - initial_trees
    return trees_planted
\end{lstlisting}
\end{analysisbox}
\end{table}
\lstset{
    comment=[l]{:},
    commentstyle=\color{blue}, 
    basicstyle=\tiny
}
\lstset{
    backgroundcolor=\color[RGB]{245,245,245},
    breaklines=true,
    breakindent=0pt,
    basicstyle=\ttfamily\small,
    frame=trbl,
    frameround = tttt,
}
\begin{table}[h]
\caption{Prompt for LLM-as-a-Judge on Reasoning tasks}\label{app:math:llm:judge}
\begin{analysisbox}
\textbf{System}:\newline
You are a helpful assistant that can identify the correct answer to the math problem.\newline
\textbf{User}:\newline
There are three choices to the same math problem. One uses natural language to answer the question, while the other two uses Python program to answer it. One directly attempts to answer the problem with coding the solution, while the last one will plan first and then write a code to answer it. Three of them can correctly answer the math problem. You need to identify which choice can correctly answer the math problem. Here is one example how to do it,
\newline
\newline
\{FEW\_SHOTS\}
\newline
\newline
Math problem: \{QUESTION\}
\newline
\newline
(A)
\newline
\{COT\_SOLUTION\}
\newline
(B)
\newline
\{PAL\_SOLUTION\}
\newline
(C)
\newline
\{P2C\_SOLUTION\}
\newline
Which of the above three choices can correctly answer the math problem?
\end{analysisbox}
\end{table}
\begin{table}[h]
\caption{Few shot Prompt for LLM as Judge on Reasoning tasks}\label{app:math:judge:shots}
\begin{analysisbox}
\textbf{User}:\newline
Math problem: There were nine computers in the server room. Five more computers were installed each day, from monday to thursday. How many computers are now in the server room?\newline
\newline
(A)\newline
Answer:\newline
There were originally 9 computers.\newline
For each of 4 days from monday to thursday, 5 more computers were added, that is, total 5 computers added for 4 days.\newline
So there are 9 + 5 = 14 computers now.\newline
So the answer is 14.\newline
\newline
(B)\newline
\begin{lstlisting}
def solution():
    computers_initial = 9
    computers_per_day = 5
    num_days = 4
    computers_added = computers_per_day * num_days
    computers_total = computers_initial + computers_added
    result = computers_total
    return result
\end{lstlisting}
(C)
\begin{lstlisting}
def solution():
    '''
    Create a function solution that returns the answer of the following question: There were nine computers in the server room. Five more computers were installed each day, from monday to thursday. How many computers are now in the server room?
    
    Let's think step by step.
    1. Calculate the total number of computers installed from Monday to Thursday: 5 computers/day * 3 days = 15
    2. Add the initial 9 computers to the total: 9 + 15 = 24
    3. There are now 24 computers in the server room.
    '''
    return 9 + 5 * 3
\end{lstlisting}
Which of the above three choices can correctly answer the math problem?\newline
\textbf{Assistant}:\newline
(B) can correctly answer the math problem. Because (A) fails to add up computers added each day, and (C) wrongly figured the number of days for computers to be added as 3 days.
\end{analysisbox}
\end{table}
\subsection{SVAMP}\label{app:svamp}
The SVAMP dataset~\cite{patel-etal-2021-nlp}  is a benchmark of 1,000 carefully constructed math word problems, designed to evaluate the robustness of models at solving elementary-level arithmetic tasks. By introducing variations that test question sensitivity, reasoning ability, and structural invariance. SVAMP minimizes reliance on shallow heuristics and exposes the limitations of state-of-the-art models, despite the simplicity of the problems for humans.

Table~\ref{app:tab:svamp} shows the results on the SVAMP dataset, utlizing CoT, P2C and PAL as a base strategies. In the SVAMP benchmark, Strategy Fusion demonstrated strong performance across both GPT-4o-mini and GPT-4o models, achieving scores of 94.4 and 95.2, respectively.  
Although PAL slightly outperformed Strategy Fusion in GPT-4o-mini with a score of 94.6, Strategy Fusion still showcased competitive results, narrowly trailing behind.  
For GPT-4o, PAL achieved the highest score of 96.0, but Strategy Fusion closely followed, highlighting its robustness and adaptability across diverse reasoning tasks.  
The SVAMP dataset, designed to test robustness to subtle variations in reasoning, benefits from the integrated approach of Strategy Fusion, which combines the strengths of multiple strategies to address complex problem-solving challenges effectively.
Similarly, with Gemini-1.5 flash, Strategy Fusion achieved a score of 94.2 on SVAMP, closely matching PAL’s 94.3 and outperforming the other strategies. This result reaffirms that integrating multiple complementary solutions through Strategy Fusion can maintain high accuracy levels on different model scales, demonstrating the adaptability of the technique to varied computational settings.
\begin{table}[h]
    \centering
    \caption{Accuracy rates, expressed as percentages, of the SVAMP benchmarks using various strategies and LLMs. Best results are shown in \textbf{bold} and second best are \underline{underlined}. }
    \label{app:tab:svamp}
    \footnotesize
    \resizebox{0.5\columnwidth}{!}{
    \begin{tabular}{l || c  | c  | c }
    \hline
         \textbf{Strategy} & \multicolumn{3}{c}{\textbf{SVAMP}} \\

         \hline 
         &\multicolumn{1}{c|}{\textbf{Gemini-1.5-flash}} &\multicolumn{1}{c|}{\textbf{GPT-4o-mini}} &  
         \multicolumn{1}{c}{\textbf{GPT-4o}}
         \\\hline
         \textbf{CoT} &  82.4 &  79.8& 75.8 \\
         \textbf{PAL} &  \textbf{94.3} & \textbf{94.6}& \textbf{96.0}\\
         \textbf{P2C} & 92.9 &  92.0  &  93.5 \\ \hline
         \textbf{LLM-as-a-Judge} & 81.7 &  83.1  &  77.0\\
         \textbf{SMaRT}\textsubscript{\emph{ours}} & \underline{94.2}  &  \underline{94.4}  & \underline{95.2}\\        

         \bottomrule
    \end{tabular}
    }
\end{table}

\newpage
\section{ALFWorld}\label{app:alfworld}
This section provides additional details and experimental results for the sequential decision making dataset ALFWorld. 
Subsection \ref{app:ex:alfworld_task_samples} covers examples of each of the six different types of tasks present in the dataset, along with a trajectory that solves the task. 
Subsection \ref{app:ex:alfworld_prompts} presents prompts used for all reasoning strategies presented in the paper.
Finally, subsection \ref{app:alfworld_additional_results} presents complete breakdown of Reflexion results and ablation studies.

\subsection{Task Samples}
This subsection provides examples of each type of sub-task available in the ALFWorld environment. Appendix Tables \ref{app:ex:alfworld_put_task}-\ref{app:ex:alfworld_puttwo_task} provide a randomly chosen example annotation for each of the tasks.
\label{app:ex:alfworld_task_samples}
\begin{table}[h]
\caption{ALFWorld: PUT Task}\label{app:ex:alfworld_put_task}
\begin{analysisbox}
Task: Find some apple and put it in sidetable.\\
Initial observation: You are in the middle of a room. Looking quickly around you, you see a cabinet 4, a cabinet 3, a cabinet 2, a cabinet 1, a coffeemachine 1, a countertop 1, a diningtable 3, a diningtable 2, a diningtable 1, a drawer 1, a fridge 1, a garbagecan 1, a microwave 1, a sidetable 1, a sinkbasin 1, a stoveburner 4, a stoveburner 3, a stoveburner 2, a stoveburner 1, and a toaster 1. \newline Your task is to: find some apple and put it in sidetable.\\
Action 1: \textbf{think:} To solve the task, I need to find and take an apple, then put it in sidetable.\\
Observation 1: OK.\\
Action 2: \textbf{think:} First I need to find an apple. An apple is more likely to appear in fridges (1), diningtables (1-3), sidetables (1), countertops (1), sinkbasins (1), garbagecan (1). I can check one by one, starting with fridge 1.\\
Observation 2: OK.\\
Action 3: \textbf{go to} fridge 1.\\
Observation 3: The fridge 1 is closed.\\
Action 4: open fridge 1.\\
Observation 4: You open the fridge 1. The fridge 1 is open. In it, you see a lettuce 2, a mug 2, a potato 2, and a tomato 1.\\
Exploring: ...\\
Action 11: \textbf{go to} garbagecan 1.\\
Observation 11: On the garbagecan 1, you see a apple 3, and a egg 3.\\
Action 12: \textbf{think:} Now I find an apple (3). Next, I need to take it.\\
Observation 12: OK.\\
Action 13: \textbf{take} apple 3 from garbagecan 1.\\
Observation 13: You pick up the apple 3 from the garbagecan 1.\\
Action 14: \textbf{think:} Now I find an apple (3). Next, I need to put it in/on sidetable 1.\\
Observation 14: OK.\\
Action 15: \textbf{go to} sidetable 1.\\
Observation 15: On the sidetable 1, you see a cup 1, a lettuce 1, a peppershaker 3, a potato 1, and a saltshaker 1.\\
Action 16: \textbf{put} apple 3 in/on sidetable 1.\\
Observation 16: You put the apple 3 in/on the sidetable 1.
\end{analysisbox}
\end{table}
\begin{table}[h]
\caption{ALFWorld: CLEAN Task}\label{app:ex:alfworld_clean_task}
\begin{analysisbox}
Task: Put a clean lettuce in diningtable.\\
Initial observation: You are in the middle of a room. Looking quickly around you, you see a cabinet 13, a cabinet 12, a cabinet 11, a cabinet 10, a cabinet 9, a cabinet 8, a cabinet 7, a cabinet 6, a cabinet 5, a cabinet 4, a cabinet 3, a cabinet 2, a cabinet 1, a coffeemachine 1, a countertop 1, a diningtable 1, a drawer 1, a fridge 1, a garbagecan 1, a microwave 1, a shelf 3, a shelf 2, a shelf 1, a sinkbasin 1, a stoveburner 4, a stoveburner 3, a stoveburner 2, a stoveburner 1, and a toaster 1. \newline Your task is to: put a clean lettuce in diningtable.\\
Action 1: \textbf{think:} To solve the task, I need to find and take a lettuce, then clean it with sinkbasin, then put it in diningtable.\\
Observation 1: OK.\\
Action 2: \textbf{think:} First I need to find a lettuce. A lettuce is more likely to appear in fridge (1), diningtable (1), sinkbasin (1), stoveburner (1-3), cabinet (1-13). I can check one by one, starting with fridge 1.\\
Observation 2: OK.\\
Action 3: \textbf{go to} fridge 1.\\
Observation 3: The fridge 1 is closed.\\
Action 4: open fridge 1.\\
Observation 4: You open the fridge 1. The fridge 1 is open. In it, you see a cup 3, a egg 2, a potato 3, and a potato 2.\\
Action 5: \textbf{go to} diningtable 1.\\
Observation 5: On the diningtable 1, you see a apple 1, a bread 1, a butterknife 2, a cup 2, a fork 2, a knife 2, a knife 1, a ladle 1, a lettuce 1, a mug 2, a mug 1, a pan 2, a peppershaker 1, a spatula 3, a tomato 2, and a tomato 1.\\
Action 6: \textbf{think:} Now I find a lettuce (1). Next, I need to take it.\\
Observation 6: OK.\\
Action 7: \textbf{take} lettuce 1 from diningtable 1.\\
Observation 7: You pick up the lettuce 1 from the diningtable 1.\\
Action 8: \textbf{think:} Now I take a lettuce (1). Next, I need to go to sinkbasin (1) and clean it.\\
Observation 8: OK.\\
Action 9: \textbf{go to} sinkbasin 1.\\
Observation 9: On the sinkbasin 1, you see a apple 2, a ladle 2, a spoon 1, and a tomato 3.\\
Action 10: \textbf{clean} lettuce 1 with sinkbasin 1.\\
Observation 10: You clean the lettuce 1 using the sinkbasin 1.\\
Action 11: \textbf{think:} Now I clean a lettuce (1). Next, I need to put it in/on diningtable 1.\\
Observation 11: OK.\\
Action 12: \textbf{go to} diningtable 1.\\
Observation 12: On the diningtable 1, you see a apple 1, a bread 1, a butterknife 2, a cup 2, a fork 2, a knife 2, a knife 1, a ladle 1, a mug 2, a mug 1, a pan 2, a peppershaker 1, a spatula 3, a tomato 2, and a tomato 1.\\
Action 13: \textbf{put} lettuce 1 in/on diningtable 1.\\
Observation 13: You put the lettuce 1 in/on the diningtable 1.
\end{analysisbox}
\end{table}

\begin{table}[h]
\caption{HEAT Task}\label{app:ex:alfworld_heat_task}
\begin{analysisbox}
Task: Heat some egg and put it in diningtable.\\
Initial observation: You are in the middle of a room. Looking quickly around you, you see a cabinet 10, a cabinet 9, a cabinet 8, a cabinet 7, a cabinet 6, a cabinet 5, a cabinet 4, a cabinet 3, a cabinet 2, a cabinet 1, a coffeemachine 1, a countertop 3, a countertop 2, a countertop 1, a diningtable 1, a drawer 6, a drawer 5, a drawer 4, a drawer 3, a drawer 2, a drawer 1, a fridge 1, a garbagecan 1, a microwave 1, a sinkbasin 1, a stoveburner 4, a stoveburner 3, a stoveburner 2, a stoveburner 1, and a toaster 1. \newline Your task is to: heat some egg and put it in diningtable.\\
Action 1: \textbf{think:} To solve the task, I need to find and take an egg, then heat it with microwave, then put it in diningtable.\\
Observation 1: OK.\\
Action 2: \textbf{think:} First I need to find an egg. An egg is more likely to appear in fridge (1), countertop (1-3), diningtable (1), stoveburner (1-4), toaster (1), garbagecan (1), cabinet (1-10). I can check one by one, starting with fridge 1.\\
Observation 2: OK.\\
Exploration: ...\\
Action 6: \textbf{go to} countertop 3.\\
Observation 6: On the countertop 3, you see a bowl 1, a butterknife 1, a egg 2, a kettle 2, a plate 1, a sink 1, and a spatula 2.\\
Action 7: \textbf{think:} Now I find an egg (2). Next, I need to take it.\\
Observation 7: OK.\\
Action 8: \textbf{take} egg 2 from countertop 3.\\
Observation 8: You pick up the egg 2 from the countertop 3.\\
Action 9: \textbf{think:} Now I take an egg (2). Next, I need go to a microwave (1) and heat it.\\
Observation 9: OK.\\
Action 10: \textbf{go to} microwave 1.\\
Observation 10: The microwave 1 is closed.\\
Action 11: \textbf{heat} egg 2 with microwave 1.\\
Observation 11: You heat the egg 2 using the microwave 1.\\
Action 12: \textbf{think:} Now I heat an egg (2). Next, I need to put it in/on diningtable 1.\\
Observation 12: OK.\\
Action 13: \textbf{go to} diningtable 1.\\
Observation 13: On the diningtable 1, you see a apple 2, a bread 3, a egg 1, a kettle 1, a knife 1, a mug 1, a papertowelroll 1, a peppershaker 2, a potato 1, a soapbottle 1, and a spatula 1.\\
Action 14: \textbf{put} egg 2 in/on diningtable 1.\\
Observation 14: You put the egg 2 in/on the diningtable 1.
\end{analysisbox}
\end{table}

\begin{table}[h]
\caption{ALFWorld: COOL Task}\label{app:ex:alfworld_cool_task}
\begin{analysisbox}
Task: Cool some pan and put it in stoveburner.\\
Initial observation: You are in the middle of a room. Looking quickly around you, you see a cabinet 16, a cabinet 15, a cabinet 14, a cabinet 13, a cabinet 12, a cabinet 11, a cabinet 10, a cabinet 9, a cabinet 8, a cabinet 7, a cabinet 6, a cabinet 5, a cabinet 4, a cabinet 3, a cabinet 2, a cabinet 1, a coffeemachine 1, a countertop 2, a countertop 1, a diningtable 1, a drawer 5, a drawer 4, a drawer 3, a drawer 2, a drawer 1, a fridge 1, a garbagecan 1, a microwave 1, a safe 1, a sinkbasin 1, a stoveburner 4, a stoveburner 3, a stoveburner 2, a stoveburner 1, and a toaster 1. \newline Your task is to: cool some pan and put it in stoveburner.\\
Action 1: \textbf{think:} To solve the task, I need to find and take a pan, then cool it with fridge, then put it in stoveburner.\\
Observation 1: OK.\\
Action 2: \textbf{think:} First I need to find a pan. An pan is more likely to appear in stoveburner (1-4), sinkbasin (1), diningtable (1), countertop (1-2), cabinet (1-16), drawer (1-5). I can check one by one, starting with stoveburner 1.\\
Observation 2: OK.\\
Exploration: ...\\
Action 5: \textbf{go to} stoveburner 3.\\
Observation 5: On the stoveburner 3, you see a pan 1.\\
Action 6: \textbf{take} pan 1 from stoveburner 3.\\
Action 7: \textbf{think:} Now I find a pan (1). Next, I need to take it.\\
Observation 6: OK.\\
Observation 7: You pick up the pan 1 from the stoveburner 3.\\
Action 8: \textbf{think:} Now I take a pan (1). Next, I need to go to a fridge (1) and cool it.\\
Observation 8: OK.\\
Action 9: \textbf{go to} fridge 1.\\
Observation 9: The fridge 1 is closed.\\
Action 10: \textbf{cool} pan 1 with fridge 1.\\
Observation 10: You cool the pan 1 using the fridge 1.\\
Action 11: \textbf{think:} Now I cool a pan (1). Next, I need to put it in/on stoveburner 1.\\
Observation 11: OK.\\
Action 12: \textbf{go to} stoveburner 1.\\
Observation 12: On the stoveburner 1, you see nothing.\\
Action 13: \textbf{put} pan 1 in/on stoveburner 1.\\
Observation 13: You put the pan 1 in/on the stoveburner 1.
\end{analysisbox}
\end{table}
\begin{table}[h]
\caption{ALFWorld: EXAMINE/LOOK Task}\label{app:ex:alfworld_examine/look_task}
\begin{analysisbox}
Task: Look at bowl under the desklamp.\\
Initial observation: You are in the middle of a room. Looking quickly around you, you see a bed 1, a desk 1, a drawer 3, a drawer 2, a drawer 1, a garbagecan 1, a safe 1, a shelf 5, a shelf 4, a shelf 3, a shelf 2, a shelf 1, a sidetable 2, and a sidetable 1. \newline Your task is to: look at bowl under the desklamp.\\
Action 1: \textbf{think:} To solve the task, I need to find and take a bowl, then find and use a desklamp.\\
Observation 1: OK.\\
Action 2: \textbf{think:} First I need to find a bowl. A bowl is more likely to appear in drawer (1-3), desk (1), sidetable (1-2), shelf (1-5), garbagecan (1). I can check one by one, starting with drawer 1.\\
Observation 2: OK.\\
Exploration: ...\\
Action 11: \textbf{go to} shelf 2.\\
Observation 11: On the shelf 2, you see a bowl 1.\\
Action 12: \textbf{think:} Now I find a bowl (1). Next, I need to take it.\\
Observation 12: OK.\\
Action 13: \textbf{take} bowl 1 from shelf 2.\\
Observation 13: You pick up the bowl 1 from the shelf 2.\\
Action 14: \textbf{think:} Now I take a bowl (1). Next, I need to find a desklamp. A desklamp is more likely to appear in desk (1), sidetable (1-2), shelf (1-5), bed (1), drawer (1-3). I can check one by one, starting with desk 1.\\
Observation 14: OK.\\
Action 15: \textbf{go to} desk 1.\\
Observation 15: On the desk 1, you see a alarmclock 2, a book 1, a cellphone 1, a keychain 1, a laptop 2, a laptop 1, and a pen 3.\\
Action 16: \textbf{go to} sidetable 1.\\
Observation 16: On the sidetable 1, you see a cd 1, a pen 1, and a pencil 1.\\
Action 17: \textbf{go to} sidetable 2.\\
Observation 17: On the sidetable 2, you see a alarmclock 1, a desklamp 1, and a pen 2.\\
Action 18: \textbf{think:} Now I find a desklamp (1). Next, I need to use it.\\
Observation 18: OK.\\
Action 19: use desklamp 1.\\
Observation 19: You turn on the desklamp 1.
\end{analysisbox}
\end{table}
\begin{table}[h]
\caption{ALFWorld: PUT TWO Task }\label{app:ex:alfworld_puttwo_task}
\begin{analysisbox}
Task: Put two saltshaker in drawer.\\
Initial observation: You are in the middle of a room. Looking quickly around you, you see a cabinet 20, a cabinet 19, a cabinet 18, a cabinet 17, a cabinet 16, a cabinet 15, a cabinet 14, a cabinet 13, a cabinet 12, a cabinet 11, a cabinet 10, a cabinet 9, a cabinet 8, a cabinet 7, a cabinet 6, a cabinet 5, a cabinet 4, a cabinet 3, a cabinet 2, a cabinet 1, a coffeemachine 1, a countertop 3, a countertop 2, a countertop 1, a drawer 3, a drawer 2, a drawer 1, a fridge 1, a garbagecan 1, a microwave 1, a shelf 3, a shelf 2, a shelf 1, a sinkbasin 1, a stoveburner 4, a stoveburner 3, a stoveburner 2, a stoveburner 1, and a toaster 1. \newline Your task is to: put two saltshaker in drawer.\\
Action 1: \textbf{think:} To solve the task, I need to find and take the first saltshaker, then put it in drawer, then find and take the second salkshaker, then put it in drawer.\\
Observation 1: OK.\\
Action 2: \textbf{think:} First I need to find the first saltshaker. A saltshaker is more likely to appear in drawer (1-3), countertop (1-3), shelf (1-3), cabinet (1-20). I can check one by one, starting with drawer 1.\\
Observation 2: OK.\\
Exploration: ...\\
Action 11: \textbf{go to} countertop 3.\\
Observation 11: On the countertop 3, you see a bowl 2, a houseplant 1, a knife 2, a peppershaker 1, a saltshaker 4, a soapbottle 1, a spatula 2, and a tomato 3.\\
Action 12: \textbf{think:} Now I find the first saltshaker (4). Next, I need to take it.\\
Observation 12: OK.\\
Action 13: \textbf{take} saltshaker 4 from countertop 3.\\
Observation 13: You pick up the saltshaker 4 from the countertop 3.\\
Action 14: \textbf{think:} Now I take the first saltshaker (4). Next, I need to put it in/on drawer 1.\\
Observation 14: OK.\\
Action 15: \textbf{go to} drawer 1.\\
Observation 15: The drawer 1 is open. In it, you see nothing.\\
Action 16: \textbf{put} saltshaker 4 in/on drawer 1.\\
Observation 16: You put the saltshaker 4 in/on the drawer 1.\\
Action 17: \textbf{think:} Now I put the first saltshaker in drawer. Next, I need to find the second saltshaker. I have visited drawer (1-3), countertop (1-3), so the second saltshaker is more likely to appear in shelf (1-3), cabinet (1-20). I can check one by one, starting with shelf 1.\\
Observation 17: OK.\\
Exploration: ...\\
Action 25: \textbf{go to} cabinet 3.\\
Observation 25: The cabinet 3 is closed.\\
Action 26: open cabinet 3.\\
Observation 26: You open the cabinet 3. The cabinet 3 is open. In it, you see a saltshaker 3, and a saltshaker 2.\\
Action 27: \textbf{think:} Now I find the second saltshaker (2). Next, I need to take it.\\
Observation 27: OK.\\
Action 28: \textbf{take} saltshaker 2 from cabinet 3.\\
Observation 28: You pick up the saltshaker 2 from the cabinet 3.\\
Action 29: \textbf{think:} Now I take the second saltshaker (2). Next, I need to put it in/on drawer 1.\\
Observation 29: OK.\\
Action 30: \textbf{go to} drawer 1.\\
Observation 30: The drawer 1 is open. In it, you see a saltshaker 4.\\
Action 31: \textbf{put} saltshaker 2 in/on drawer 1.\\
Observation 31: You put the saltshaker 2 in/on the drawer 1.
\end{analysisbox}
\end{table}

\subsection{ ALFWorld Prompts}
\label{app:ex:alfworld_prompts}
We use two-shot prompting for all the settings. However, for easier readability, this section demonstrates one-shot prompts. An example of one of the Pick tasks has been shared below.
The system prompt shared among all experimental settings can be found at Table \ref{app:ex:alfworld_system_prompt}. The prompts used for base strategies are presented in Table \ref{app:ex:alfworld_direct_prompt} and Table \ref{app:ex:alfworld_cot_prompt} for Direct, and CoT/ReAct strategies respectively.
Finally the LLM-as-a-Judge prompt can be located at Table \ref{app:ex:alfworld_llm_as_a_judge_prompt}.
\begin{table}[h]
\caption{ALFWorld System Prompt}\label{app:ex:alfworld_system_prompt}

\begin{analysisbox}
Interact with a household to solve a task.

Do not think more than thrice consecutively.

You need to generate actions that strictly follow the below templates:\\
1. goto [location] \\
2. take [object] from [location] put [object] in/on [location]\\
3. open [something]\\
4. close [something]\\
5. toggle [object][location]\\
6. clean [object] with [something]\\
7. heat [object] with [receptacle]\\
8. cool [object] with [receptacle]\\

Do not include any additional text apart from the action based on above templates.\\
Remember you need to go to a location or device before using it.\\
Once you are at a microwave, you can directly try to heat the item.\\
For example: The action `go to microwave 1' can directly be followed by the action `heat apple 1 with microwave 1'\\
Once you are at a fridge, you can directly try to cool the item.\\
For example: The action `go to fridge 1' can directly be followed by the action `cool bread 1 with fridge 1'
\end{analysisbox}
\end{table}
\begin{table}[h]
\caption{ALFWorld Direct/Act Prompt}\label{app:ex:alfworld_direct_prompt}
\begin{analysisbox}
Here are two examples. They are very relevant. Please use the actions in these examples as your guidelines.\\
You are in the middle of a room. Looking quickly around you, you see a coffeetable 1, a diningtable 1, a drawer 4, a drawer 3, a drawer 2, a drawer 1, a dresser 1, a garbagecan 1, a sidetable 2, a sidetable 1, and a sofa 1.\\
Your task is to: put two cellphone in sofa.\\
> go to coffeetable 1\\
On the coffeetable 1, you see a box 1, a cellphone 3, and a plate 1.\\
> take cellphone 3 from coffeetable 1\\
You pick up the cellphone 3 from the coffeetable 1.\\
> go to sofa 1\\
On the sofa 1, you see a newspaper 2, a pillow 1, a remotecontrol 2, and a remotecontrol 1.\\
> put cellphone 3 in/on sofa 1\\
You put the cellphone 3 in/on the sofa 1.\\
> go to diningtable 1\\
On the diningtable 1, you see a cellphone 2, a keychain 2, a laptop 1, a statue 2, and a statue 1.\\
> take cellphone 2 from diningtable 1\\
You pick up the cellphone 2 from the diningtable 1.\\
> go to sofa 1\\
On the sofa 1, you see a cellphone 3, a newspaper 2, a pillow 1, a remotecontrol 2, and a remotecontrol 1.\\
> put cellphone 2 in/on sofa 1\\
You put the cellphone 2 in/on the sofa 1.\\

\textit{Example 2: Truncated}
\end{analysisbox}
\end{table}

\begin{table}[h]
\caption{ALfWorld CoT/ReAct Prompt}\label{app:ex:alfworld_cot_prompt}
\begin{analysisbox}
Here are two examples. They are very relevant. Please use the actions in these examples as your guidelines.\\ 
You are in the middle of a room. Looking quickly around you, you see a coffeetable 1, a diningtable 1, a drawer 4, a drawer 3, a drawer 2, a drawer 1, a dresser 1, a garbagecan 1, a sidetable 2, a sidetable 1, and a sofa 1.\\ 
Your task is to: put two cellphone in sofa.\\ 
\textbf{> think:} To solve the task, I need to find and take the first cellphone, then put it in sofa, then find and take the second cellphone, then put it in sofa.\\ 
OK.\\ 
\textbf{> think:} First I need to find the first cellphone. A cellphone is more likely to appear in coffeetable (1), diningtable (1), sidetable (1-2), drawer (1-4), sofa (1), dresser (1), garbagecan (1). I can check one by one, starting with coffeetable 1.\\ 
OK.\\ 
> go to coffeetable 1\\ 
On the coffeetable 1, you see a box 1, a cellphone 3, and a plate 1.\\ 
> take cellphone 3 from coffeetable 1\\ 
You pick up the cellphone 3 from the coffeetable 1.\\ 
\textbf{> think:} Now I find the first cellphone (3). Next, I need to put it in/on sofa 1.\\ 
OK.\\ 
> go to sofa 1\\ 
On the sofa 1, you see a newspaper 2, a pillow 1, a remotecontrol 2, and a remotecontrol 1.\\ 
> put cellphone 3 in/on sofa 1\\ 
You put the cellphone 3 in/on the sofa 1.\\ 
\textbf{> think:} Now I put the first cellphone in sofa. Next, I need to find the second cellphone. I have visited coffeetable (1), so the second cellphone is more likely to appear in diningtable (1), sidetable (1-2), drawer (1-4), sofa (1), dresser (1), garbagecan (1). I can check one by one, starting with coffeetable 1.\\ 
OK.\\ 
> go to diningtable 1\\ 
On the diningtable 1, you see a cellphone 2, a keychain 2, a laptop 1, a statue 2, and a statue 1.\\ 
\textbf{> think:} Now I find the second cellphone (2). Next, I need to take it.\\ 
OK.\\ 
> take cellphone 2 from diningtable 1\\ 
You pick up the cellphone 2 from the diningtable 1.\\ 
\textbf{> think:} Now I take the second cellphone (2). Next, I need to put it in/on sofa 1.\\ 
OK.\\ 
> go to sofa 1\\ 
On the sofa 1, you see a cellphone 3, a newspaper 2, a pillow 1, a remotecontrol 2, and a remotecontrol 1.\\ 
> put cellphone 2 in/on sofa 1\\ 
You put the cellphone 2 in/on the sofa 1.\\ 
\\

\textit{Example 2: Truncated}
\end{analysisbox}
\end{table}

\begin{table}[h]
\caption{ALFWorld LLM-as-a-Judge Prompt}\label{app:ex:alfworld_llm_as_a_judge_prompt}
\begin{analysisbox}
You are an evaluator and should choose one of these plans to solve the following goal. Return the plan number and no other text.\\

Here is the information about the task: ...\\

** Plan 1: **\\
Action 1: ...\\
Observation 1: ...\\
... \\

** Plan 2: **\\
Action 1: ...\\
Observation 1: ...\\
...\\

The correct plan is Plan:
\end{analysisbox}
\end{table}
SMaRT process involves upto 2 calls. 
During the fusion phase the LLM first generates an exploration report consisting of all the locations visited by the LLM during the base strategies and the objects found at respective locations. 
Then ReAct approach is used with the augmentation of environment summary and failed base trajectories, to generate a new trajectory. Refer to Tables~\ref{app:ex:alfworld_sff_gen_env_info} and \ref{app:ex:alfworld_sff_gen_new_plan}.




\begin{table}[h]
\label{app:ex:alfworld_gen_env_info}
\caption{ALFWorld - Generate Environment Information}\label{app:ex:alfworld_sff_gen_env_info}
\begin{analysisbox}
Task Description:\{...\}\\\\
Trajectory Option 1:\\
Action 1: ...\\
Observation 1: ...\\
...\\

Trajectory Option 2:\\
Action 1: ...\\
Observation 1: ...\\
...\\
\\
Return a list of unique locations with objects found at these locations based on the trajectories provided. You should return in the format of "Location j: Object k" only.
\end{analysisbox}
\end{table}
\begin{table}[h]
\caption{ALFWorld SMaRT Prompt - ReAct +  Generated Environment Information}
\label{app:ex:alfworld_sff_gen_new_plan}
\begin{analysisbox}
You should avoid any actions already tried in the failed trajectory options which are invalid or irrelevant to the task.\\
Failed Trajectory Option 1:\\
Action 1: ...\\
Observation 1: ...\\
\\
Failed Trajectory Option 2:\\
Action 1: ...\\
Observation 1: ...\\
\\
Here is the information about the environment.: `Output of the LLM response based on Appendix Table \ref{app:ex:alfworld_gen_env_info}' prompt.
\end{analysisbox}
\end{table}
\subsection{ALFWorld: Additional Results}
\label{app:alfworld_additional_results}
\textbf{Reflexion Baselines:} Table~\ref{tab:alf_Reflexion} displays the success rate of ALFWorld when employing the Reflexion strategy. We used the code available on their repository \footnote{https://github.com/noahshinn/reflexion} without making any changes. Please note that the original paper uses GPT-4 model with upto 15 trials and obtains > 90\% success rate. We have truncated the number of trials to ensure fair comparison with our approach.
\begin{table}[h]
    \centering
    \caption{The success rate of the ALFWorld benchmark using the Reflection strategy.}
    \label{tab:alf_Reflexion}
    \resizebox{0.7\columnwidth}{!}{
    \begin{tabular}{l || c | c | c | c | c | c | c }
    \toprule
    \multicolumn{8}{c}{\textbf{ALFWorld environment}} \\ \hline 
     & \textbf{Pick} & \textbf{Clean} & \textbf{Heat} & \textbf{Cool} & \textbf{Look} & \textbf{Pick2} & \textbf{All} \\ \hline\hline 
    
    \multicolumn{8}{c}{\textbf{Gemini-1.5}} \\\hline
    \textbf{Reflexion} (5 Trials) & 54 & 38 & 21 & 42 & 50 & 17 & 38.06 \\ \hline
       
    \multicolumn{8}{c}{\textbf{GPT-4o}} \\ \hline    
    \textbf{Reflexion} (5 Trials) & 91 & 74 & 8 & 85 & 33 & 82 & 63.43 \\ \hline
     \bottomrule
        \end{tabular}
        }
\end{table}
\\
\textbf{Impact of Environment Summarization:} We use an additional call in ALFWorld setting to summarize the environment observations from over 60 LLM calls (60 action and observation pairs). Specifically, each base strategy is limited to 30 turns to complete the task, which leads to over 60 lines of information per strategy comprising of action and observation pairs. 
In order to reduce LLM hallucination mistakes, we have added a summary section, alongside of the base trajectories. 
The same LLM used for response generation is used for environment summary generation as well.
The inclusion of the environment feedback information leads to a success rate of 83.58\% and 96.27\% for Gemini 1.5 Flash and GPT-4o respectively.
In an ablation study we noticed that even without the environment information, we obtain the best performance with Gemini 1.5 Flash obtaining 79.85\% and GPT-4o obtaining 87.31\%, registering a 4-9 \% performance drop.
Due to this reason, we presented the results of experiments containing the environment summary which does not impact the core SMaRT functionality itself.

\newpage
\section{Natural Plan}\label{app:natural_plan}
This benchmark was designed to evaluate the planning capabilities of LLMs when tasks are expressed in natural language.
It focuses on three distinct planning categories: Trip Planning, Meeting Planning, and Calendar Scheduling.
To decouple the use of external tools and isolate the planning capabilities of LLMs, the dataset was curated using resources such as Google Flights, Google Maps, and Google Calendar by the authors of \cite{zheng2024natural}.
Each category includes task instructions along with reference information necessary to complete the task and five illustrations corresponding to each data-point. 
Below, we provide example task-solution pairs for each category, along with the prompts used for different strategies.

To assess performance across the various task categories, the dataset includes golden truth plan for each data-point, serving as a benchmark for evaluation.
For Meeting Planning and Trip Planning, the evaluation metric is the Exact Match (EM) score, which measures whether the LLM-generated plan aligns perfectly with the golden plan. This ensures an objective and consistent evaluation of task performance.
Calendar Scheduling tasks involve additional complexity due to participant availability constraints. 
Alongside the EM score, evaluation considers cases where common timeslot exists among participants. 
In such scenarios, the evaluation includes assessing whether the model identifies alternative scheduling solutions that satisfy the required duration as specified in the task query.
Using the specified prompts and the described evaluation criteria~(Refer to Tables~\ref{app:ex:natural_plan_trip_planning} to~\ref{app:ex:natural_plan_prompt_structure} for exact prompts), we computed the success rates for tasks across varying complexities. These complexities are defined by the number of cities in Trip Planning, the number of participants in Meeting Planning and the number of attendees or scheduling days in Calendar Scheduling.

The results are detailed in Tables \ref{app:tab:natural_plan_trip_planning}, \ref{app:tab:natural_plan_meeting_planning}, and~\ref{app:tab:natural_plan_calendar_scheduling}. 
These tables illustrate how model performance varies as the complexity of the task query increases.
\begin{table}[h]
\caption{Trip planning}\label{app:ex:natural_plan_trip_planning}
\begin{analysisbox}
\textbf{Task:}\\You plan to visit 3 European cities for 7 days in total.\\
 You only take direct flights to commute between cities.\\
 You want to spend 4 days in Berlin.\\
 You would like to visit Venice for 3 days.\\
 You want to spend 2 days in Tallinn.\\
 Find a trip plan of visiting the cities for 7 days by taking direct flights to commute between them.\\

\textbf{Reference information:}\\
* Special Events: You are going to attend a wedding in Venice on day 5 and day 7.\\
* Direct Flights: Tallinn to Berlin, Berlin to Venice\\
     
\textbf{Solution:} Here is the trip plan for visiting the 3 European cities for 7 days: \\
**Day 1-2:** Arriving in Tallinn and visit Tallinn for 2 days. \\
**Day 2:** Fly from Tallinn to Berlin.\\
**Day 2-5:** Visit Berlin for 4 days.\\
**Day 5:** Fly from Berlin to Venice.\\
**Day 5-7:** Visit Venice for 3 days.\\

\end{analysisbox}
\end{table}
\begin{table}[h]
\caption{Meeting planning}
\begin{analysisbox}

\textbf{Task:} \\ You are visiting San Francisco for the day and want to meet as many friends as possible. Solve the problem by considering various different schedules and picking the best one to optimize your goals. \\

\textbf{Reference information:} Travel distances (in minutes): \\
Haight-Ashbury to Chinatown: 19. \\
Haight-Ashbury to Union Square: 17. \\
Chinatown to Haight-Ashbury: 19. \\
Chinatown to Union Square: 7. \\
Union Square to Haight-Ashbury: 18. \\
Union Square to Chinatown: 7. \\

You arrive at Haight-Ashbury at 9:00AM \\
Kimberly will be at Chinatown from 2:45PM to 6:15PM \\
You'd like to meet Kimberly for a minimum of 120 minutes \\
Ronald will be at Union Square from 2:00PM to 9:30PM \\
You'd like to meet Ronald for a minimum of 30 minutes \\

\textbf{Solution:} \\ 
You start at Haight-Ashbury at 9:00AM \\
You travel to Chinatown in 19 minutes and arrive at 9:19AM \\
You wait until 2:45PM \\
You meet Kimberly for 120 minutes from 2:45PM to 4:45PM \\
You travel to Union Square in 7 minutes and arrive at 4:52PM \\
You meet Ronald for 30 minutes from 4:52PM to 5:22PM

\end{analysisbox}
\end{table}
\begin{table}[h]
\caption{Calandar Scheduling}
\begin{analysisbox}

\textbf{Task:} \\ You need to schedule a meeting for Roger, Karen and Dorothy for half an hour between the work hours of 9:00 to 17:00 on Monday. \\

\textbf{Reference information:} \\ Here are the existing schedules for everyone during the day: \\
Roger's calendar is wide open the entire day. \\
Karen has meetings on Monday during 10:00 to 10:30, 11:30 to 12:00, 12:30 to 13:00, 14:00 to 15:00, 15:30 to 16:00; \\
Dorothy is busy on Monday during 9:00 to 10:00, 10:30 to 11:00, 11:30 to 12:00, 12:30 to 13:00, 14:00 to 15:30, 16:00 to 17:00;\\
\\
You would like to schedule the meeting at their earlist availability.\\
Find a time that works for everyone's schedule and constraints. \\

\textbf{Solution:} \\
Here is the proposed time: Monday, 11:00 - 11:30
\end{analysisbox}
\end{table}
\begin{table}[h]
\caption{Prompts for Direct strategy in Natural Plan}
\begin{analysisbox}
\label{app:ex:natural_plan_direct_prompts}

\textbf{BASIC\_MEETING\_TASK\_INSTRUCTIONS:}\\
You are a proficient meeting planner.\\
You are provided with a Query, Reference Information and illustrations of Meeting Plan. \\
Using the provided Reference Information and Query, please give me a detailed Meeting Plan.\\
Only return the final plan for query by strictly following the same plan format as shown in illustrations.\\

\textbf{BASIC\_SCHEDULING\_TASK\_INSTRUCTIONS:}\\
You are an expert at scheduling meetings. \\
You are given a few constraints on the existing schedule of each participant, the meeting duration, and possibly some preferences on the meeting time. \\
Note there exists a solution that works with existing schedule of every participant. \\
Once you’ve finalized the schedule, present it in this format: SOLUTION: Here is the proposed time: [Insert Time]. \\

\textbf{BASIC\_TRIP\_TASK\_INSTRUCTIONS:}\\
You are an expert at planning trips.\\
You are given a few constraints regarding the cities to visit and the durations of staying at each city along with direct flight routes.\\
Return the final plan for query by strictly following the same plan format as shown in illustrations.\\
\end{analysisbox}
\end{table}
\begin{table}[h]
\caption{Prompts for CoT strategy in Natural Plan }
\begin{analysisbox}
\label{app:ex:natural_plan_cot_prompts}
\textbf{STEP\_BY\_STEP\_THINKING\_FOR\_MEETING\_PLANNING:} Think step-by-step to create a final meeting planning:\\
    * Step 1: Understand Key Constraints and Objective:\\
        - Objective: Meet as many friends as possible.\\
        - Constraints: Arrival time, required meeting time with friends, and travel distances.\\
    * Step 2: Identify Key Time Blocks: \\
        - Determine when each friend is available and how much time you want to spend with them. \\
        - Allow potential meetings with other friends.\\
    * Step 3: Account for Travel Times\\
        - Calculate travel time between different locations to plan when you can arrive and how much buffer time you have. \\
        - You can wait or travel to another friend, considering time availability.\\
    * Step 4: Optimize Schedule for Efficiency\\
        - After calculating all travel and wait times, adjust your meeting schedule to minimize idle time.\\
        - Ensure that travel times do not prevent you from fulfilling the minimum meeting time requirement.\\
    * Step 5: Build the Plan: Create the meeting plan in the required format using only the provided reference data with no repetition of meeting same people\\

\textbf{STEP\_BY\_STEP\_THINKING\_FOR\_CALENDAR\_SCHEDULING:} Think step-by-step to create a final calendar schedule and explain your reasoning:\\
    * Step 1: Understand Key Constraints and Objective:\\
        - Objective: Schedule a meeting that fits within everyone's availability.\\
        - Constraints: Meeting duration (30 minutes or 1 hour), work hours (9:00 to 17:00), attendee availability, and blocked time for existing meetings.\\
    * Step 2: Identify Available Time Blocks for All Attendees:\\
        - Examine each attendee’s calendar to find when they are available during the specified day(s).\\
        - Eliminate times that overlap with existing meetings or blocked time.\\
    * Step 3: Optimize for the Best Meeting Time:\\
        - Select the available time slot that fits all attendees’ schedules and avoids overlap with existing meetings.\\
        - If there are multiple options, prioritize the time that minimizes idle time or gaps in schedules.\\
    * Step 4: Build the Final Schedule:\\
        - Create the final calendar schedule, clearly indicating the day and time of the meeting.\\
        - Ensure the format aligns with the reference structure and adheres to all constraints. Final plan should be given as ``SOLUTION: Here is the proposed time:''\\

\textbf{STEP\_BY\_STEP\_THINKING\_FOR\_TRIP\_PLANNING:}Think step-by-step to create a final trip plan and explain your reasoning:\\
    * Step 1: Understand Key Constraints and Objective\\
        - Objective: Plan a trip visiting multiple cities for a specified number of days.\\
        - Constraints: Such as Total number of days for the trip, Duration of stay in each city, Direct flight connections and Specific events (e.g., attending a wedding) on certain days.\\
    * Step 2: Identify City Visit Durations\\
        - Ensure the total number of days across cities matches the overall trip duration.\\
        - Pay attention to specific constraints like fixed events (e.g., attending a wedding) that require being in a city on certain days.\\
    * Step 3: Consider Direct Flight Routes only\\
        - Examine the flight connections provided to understand which cities can be directly reached from one another.\\
        - If no direct flight exists between two cities, you cannot travel directly between them.\\
    * Step 4: Build the Final Trip Plan\\
        - Create the trip plan using the required format with day-by-day breakdowns.\\
        - Ensure the format aligns with the reference structure and adheres to all constraints.\\
\end{analysisbox}
\end{table}

\begin{table}[h]
\caption{Prompt for LLM-as-a-Judge strategy~(for all categories) in Natural Plan}\label{app:ex:natural_plan_judge_prompt}

\begin{analysisbox}

\textbf{LLM-AS-A-JUDGE\_PROMPT:}\\You are a proficient planner.\\
You are provided with a Query, Reference Information and two Plans.\\
Provided a Query and two different plans - (A) and (B), you need to select the best plan which fulfills most of the constraints for the task. \\
You have to select the best plan by providing the index of the plan. \\
For example:\\
    * If the first plan (Plan (A)) is the best, respond with index (A)\\
    * If the second plan (Plan (B)) is the best, respond with index (B)\\

\end{analysisbox}
\end{table}

\begin{table}[h]
\caption{Prompt for Strategy Fusion in Natural Plan~(Meeting Planning)}\label{app:ex:natural_plan_sff_meeting_plan_prompt}

\begin{analysisbox}

\textbf{STRATEGY\_FUSION\_INSTRUCTION\_MEETING\_PLANNING:} You are a proficient planning assistant capable of creating an optimized meeting plan. \\
Given two versions of a plan, your task is to create a final plan that best fulfills all the constraints of the query.\\
Make sure that each sentence in final plan should begin with either of prefix only - You start, You wait, You meet or You travel.\\
\\
Task Details:\\
    * You are provided with a Query, Reference Information, and two Plans (Plan A and Plan B).\\
    * You may cross-reference and mix various components from both plans and validate or use reference information to create a final plan.\\
    * Your goal is to optimize the schedule, ensuring it satisfies all constraints are met properly.\\
    * Make sure that each sentence in final plan should begin with either of prefix only - You start, You wait, You meet or You travel.\\
\\
Instructions:\\
    * Prioritize meeting the required time with each friend, accounting for their availability and travel times.\\
    * Make sure to pay keen attention to the start and end time of meeting people. You must double check before crafting final plan.\\
    * You must consider all constraints (e.g., arrival times, meeting durations, and travel distances) when crafting the plan.\\
    * Carefully evaluate each solution on key constraints: travel times, meeting durations, and availability, and identify the strongest elements.\\
    * Account for any wait times in plan if availability starts after your arrival and avoid backtracking. Do not go back in time.\\
    * Combine these elements to create a new solution that best optimizes the query’s goals to ensure the final plan is cohesive and satisfies all requirements. \\
    * Ensure all times follow a logical sequence without referencing events before the start or travel time. \\
    * Note that you may not have to wait in some cases, and you should consider this in your final plan.\\
    * Ensure that your final plan follows the same format as the provided plans.
\end{analysisbox}
\end{table}

\begin{table}[h]
\caption{Prompt for Strategy Fusion in Natural Plan~(Calendar Scheduling)}\label{app:ex:natural_plan_sff_calendar_scheduling_prompt}
\begin{analysisbox}
\textbf{STRATEGY\_FUSION\_INSTRUCTION\_CALENDAR\_SCHEDULING:} You are a proficient scheduling assistant tasked with creating an optimized calendar plan. \\
Your goal is to minimize overlaps with existing meetings and ensure smooth transitions between attendees’ schedules.\\
\\
Task Overview:\\
* You are provided with a query, reference information, and two proposed schedules for meetings with multiple attendees.\\
* Your role is to cross-reference all attendees' availability and propose the best time slot that avoids conflicts, fits the meeting duration, and optimizes time usage.\\
\\
Instructions:\\
* Check Availability: Identify time blocks that accommodate all attendees within the specified work hours, avoiding conflicts with existing meetings or blocked time.\\
* Avoid Overlaps: Ensure the selected time has no full or partial overlaps with other meetings or blocked periods.\\
* Meeting Duration: Confirm that the chosen time block supports either a 30-minute or 1-hour meeting.\\
* Cross-reference Suggestions: Compare the proposed schedules. \\
If either of suggested the schedules work for all people, copy paste the same in response.
If neither works, suggest a new time that meets all constraints (if needed).\\
* Optimize Time Usage: Prioritize the earliest available slot and minimize idle time for attendees by clustering meetings where possible.\\
* Schedule the meeting within the provided work hours (24-hour format).\\
* Avoid scheduling during busy or blocked periods for attendees.\\
* Ensure the meeting duration fits fully without any overlap or conflicts.\\
Once you finalize the schedule, present it as follows: SOLUTION: Here is the proposed time: [Insert Time].
\end{analysisbox}
\end{table}
\begin{table}[h]
\caption{Prompt for Strategy Fusion in Natural Plan~(Trip Planning)}
\label{app:ex:natural_plan_sff_trip_planning_prompt}
\begin{analysisbox}

\textbf{STRATEGY\_FUSION\_INSTRUCTION\_TRIP\_PLANNING:} You are an expert at planning trips.\\
You are provided with a Query, Reference Information (flight routes and Trip Constraints as city durations and special events), and two Plans (Plan A and Plan B).\\
Your task is to create an optimized trip plan visiting multiple cities.\\
Adhere to all the provided constraints by using information from either of the plans or referring to given information.\\
Make sure it strictly follows the format provided in the Plans.\\
\\
Instructions:
    * Plan a trip visiting multiple cities within a specified number of days.\\
    * Ensure the sum of days across all cities matches the total trip duration.\\
    * Calculate the days spent in each city to assure, they match with constraints. Duration of visit is calculated as "end\_day - start\_day + 1"\\
    * The day of flight, should be counted as day of visit to both the cities of arrival and departure. Adjust the day count accordingly.\\
    * You must attend specific events requiring you to be in certain cities on certain days (e.g., attending a wedding or meeting someone).\\
    * Only cities with direct flights between them can be visited in succession. If there’s no direct flight, the cities cannot be linked sequentially.\\
    * Use the information from two plans provided or create new plan details if something is incorrect. \\
    * Ensure that your final plan follows the same format as the provided plans.
\end{analysisbox}
\end{table}
\begin{table}[h]
\caption{Prompt Structure for various strategies in Natural Plan}\label{app:ex:natural_plan_prompt_structure}
\begin{analysisbox}

\textbf{DIRECT\_PROMPT\_FOR\_TASK:} \\
\{BASIC\_TASK\_INSTRUCTIONS\} \\
\#\# Query \\
\{Task\}\\
\#\# Reference information\\
\{reference\_information\}\\
\\
\{few-shot illustrations\}\\
\\
\textbf{COT\_PROMPT\_FOR\_TASK:} \\
\{BASIC\_TASK\_INSTRUCTIONS\}\\
\{STEP\_BY\_STEP\_THINKING\_FOR\_TASK\}\\
\#\# Query\\
\{query\}\\
\#\# Reference information\\
\{reference\_information\}\\
\\
\{few-shot illustrations\}\\
\\
\\
\textbf{LLM-AS-A-JUDGE\_PROMPT\_FOR\_TASK:} \\
\{LLM-AS-A-JUDGE\_PROMPT\}\\
\#\# Query\\
\{query\}\\
\#\# Reference information\\
\{reference\_information\}\\
\#\# Plans\\
Plan (A): \\
\{plan\_A\}\\
\\
Plan (B):\\ 
\{plan\_B\}\\
\\
\textbf{STRATEGY\_FUSION\_PROMPT\_FOR\_TASK:} \\
\{STRATEGY\_FUSION\_INSTRUCTION\_FOR\_TASK\}\\
\#\# Query\\
\{query\}\\
\#\# Reference information\\
\{reference\_information\}\\
\#\# Plans\\
Plan (A): \\
\{plan\_A\}\\
\\
Plan (B):\\ 
\{plan\_B\}
\end{analysisbox}
\end{table}

\begin{table}[h]
        \centering
        \caption{Success rates of trip planning task under various query complexity.}
        \label{app:tab:natural_plan_trip_planning}
        \begin{tabular}{l||c | ccccccccc}
            \hline
             & &\multicolumn{8}{c}{Number of Cities}\\
            \cline{3-10}
            & &  3 & 4 & 5 & 6 & 7 & 8 &9&10\\
            \hline
            \textbf{Strategy}& \textbf{Success Rate} & &  &  &  &  &  &  &\\\hline \hline
            \multicolumn{10}{c}{\textbf{Gemini-1.5}}   \\\hline
            \textbf{Direct} &32.19 &	0.69&	0.64&	0.52&	0.32&	0.18&	0.14&	0.02&	0.05\\
            \textbf{CoT} & 31.19&	0.62&	0.57&	0.52&	0.36&	0.18&	0.14&	0.05&	0.04\\ \hline
            \textbf{LLM-as-a-Judge} &29.44&	0.64&	0.61&	0.47&	0.30&	0.15&	0.12&	0.02&	0.04\\
            \textbf{Strategy Fusion}$_{\text{ours}}$ & 33.25 & 0.70 & 0.62 & 0.56 & 0.38 & 0.19 & 0.12 & 0.04 & 0.05 \\ \hline \hline

            \multicolumn{10}{c}{\textbf{GPT-4}}   \\\hline
            \textbf{Direct} & 5.56&	0.23&	0.20&	0.00&	0.01&	0.00&	0.00&	0.00&	0.00\\
            \textbf{CoT} &   9.38&	0.38&	0.32&	0.02&	0.02&	0.00&	0.00&	0.00&	0.00\\ \hline
            \textbf{LLM-as-a-Judge} & 6.25&	0.25&	0.24&	0.00&	0.02&	0.00&	0.00&	0.00&	0.00\\
            \textbf{Strategy Fusion}$_{\text{ours}}$ &24.5&	0.70&	0.68&	0.36&	0.11&	0.06&	0.03&	0.01&	0.00\\
            \hline
        \end{tabular}
    \end{table}

\begin{table}[h]
        \centering
        \caption{Success rates of meeting planning task under various query complexity.}
        \label{app:tab:natural_plan_meeting_planning}
        \begin{tabular}{l || c | ccccccccccc}
            \hline
             & &\multicolumn{10}{c}{Days}\\
            \cline{3-12}
            & & 1 & 2 & 3 & 4 & 5 & 6 & 7 & 8 &9&10\\
            \hline
            \textbf{Strategy}& \textbf{Success Rate} & &  &  &  &  &  &  &\\\hline \hline
            \multicolumn{12}{c}{\textbf{Gemini-1.5}}   \\\hline
            \textbf{Direct}& 25.2&	0.84&	0.66&	0.43&	0.30&	0.19&	0.04&	0.05&	0.01&	0.00&	0.00\\
            \textbf{CoT}&26.0&	0.81&	0.73&	0.49&	0.34&	0.15&	0.03&	0.02&	0.01&	0.01&	0.01\\ \hline
            \textbf{LLM-as-a-Judge}&26.2&	0.82&	0.69&	0.47&	0.35&	0.17&	0.05&	0.06&	0.01&	0.00&	0.00\\
            \textbf{Strategy Fusion}$_{\text{ours}}$ &27.8&	0.89&	0.74&	0.53&	0.35&	0.16&	0.06&	0.04&	0.01&	0.00&	0.00\\
            \hline \hline

            \multicolumn{12}{c}{\textbf{GPT-4}}   \\\hline
            \textbf{Direct}& 46.3&	0.98&	0.91&	0.69&	0.72&	0.50&	0.37&	0.18&	0.15&	0.06&	0.07\\
            \textbf{CoT}& 46.5&	0.95&	0.95&	0.73&	0.75&	0.44&	0.36&	0.25&	0.09&	0.10&	0.03\\ \hline
            \textbf{LLM-as-a-Judge}&45.1&	0.96&	0.95&	0.70&	0.72&	0.44&	0.33&	0.21&	0.10&	0.07&	0.03\\
            \textbf{Strategy Fusion}$_{\text{ours}}$& 49.8&	0.98&	0.95&	0.75&	0.77&	0.53&	0.43&	0.24&	0.18&	0.07&	0.08\\
            \hline
        \end{tabular}
    \end{table}

\begin{table}[h]
        \centering
        \caption{Success rates of calendar scheduling task under various query complexity.}
        \label{app:tab:natural_plan_calendar_scheduling}
        \begin{tabular}{l||c | ccccccccccc}
            \hline
             & &\multicolumn{10}{c}{Number of People (Number of Days)}\\
            \cline{3-12}
            & & 2 (1) & 2 (2) & 2 (3) & 2 (4) & 2 (5) & 3 (1) & 4 (1) & 5 (1) &6 (1) & 7(1)\\
            \hline
            \textbf{Strategy}& \textbf{Success Rate} & &  &  &  &  &  &  &\\\hline \hline
            \multicolumn{12}{c}{\textbf{Gemini-1.5}}   \\\hline
            \textbf{Direct}&50.2&	0.77&	0.68&	0.67&	0.50&	0.50&	0.64&	0.34&	0.27&	0.28&	0.37\\
            \textbf{CoT}& 57.6&	0.83&	0.72&	0.61&	0.54&	0.61&	0.65&	0.53&	0.46&	0.43&	0.38\\ \hline
            \textbf{LLM-as-a-Judge}&57.0&	0.83&	0.69&	0.63&	0.52&	0.62&	0.71&	0.52&	0.43&	0.41&	0.34\\
            \textbf{Strategy Fusion}$_{\text{ours}}$& 58.5&	0.84&	0.71&	0.63&	0.53&	0.64&	0.73&	0.53&	0.42&	0.43&	0.39\\
            \hline \hline
            
            \multicolumn{12}{c}{\textbf{GPT-4}}   \\ \hline      
            \textbf{Direct}&66.1&	0.86&	0.83&	0.85&	0.73&	0.82&	0.79&	0.44&	0.46&	0.37&	0.46\\
            \textbf{CoT}&70.2&	0.93&	0.86&	0.84&	0.76&	0.75&	0.86&	0.57&	0.57&	0.40&	0.48\\ \hline
            \textbf{LLM-as-a-Judge}&69.4&	0.95&	0.89&	0.83&	0.79&	0.79&	0.83&	0.54&	0.51&	0.35&	0.46\\
            \textbf{Strategy Fusion}$_{\text{ours}}$& 72.0&	0.94&	0.91&	0.86&	0.85&	0.79&	0.85&	0.56&	0.57&	0.42&	0.45\\
            \hline
        \end{tabular}
    \end{table}

\newpage
\section{TravelPlanner}\label{app:TravelPlanner}

The TravelPlanner benchmark~\cite{xie2024travelplanner} is designed to generate comprehensive travel plans based on user-provided textual queries. 
It offers a rich and complex environment for testing the capabilities of LLMs as agents tasked with fulfilling multiple constraints while creating detailed travel itineraries.
The dataset incorporates a variety of constraints, including both commonsense constraints and hard constraints (refer to Table 1 of~\cite{xie2024travelplanner} for detailed description of each constraint). 
While TravelPlanner is intended to evaluate the overall capabilities of agents in both tool use and planning, our focus in this study was specifically on assessing planning skills in isolation (referred to as the sole-planning mode).
To evaluate the quality of travel plans generated by LLM agents, we employed well-established performance indicators. 
These indicators provide baseline metrics to measure the LLM{'}s effectiveness in planning multi-day itineraries, enabling a robust assessment of their planning proficiency.
Indicators used are listed below:
\begin{itemize}[noitemsep,topsep=6pt,parsep=0pt,partopsep=0pt]
\item \textbf{Delivery Rate}: Evaluates if the agent can deliver a plan within 30 steps
\item \textbf{Commonsense Constraint Pass Rate}: Measures if the agent incorporates commonsense (across eight dimensions) into the plans
\item \textbf{Hard Constraint Pass Rate}: Checks if the agent meets the hard requirements specified in the query
\item \textbf{Final Pass Rate}: The proportion of plans that satisfy all the above indicators
\end{itemize}

Following the original paper, for evaluating constraint pass rates, we employed two distinct strategies: micro and macro evaluation. 
The micro evaluation computes the ratio of constraints successfully passed to the total number of constraints across all plans. 
In contrast, the macro evaluation calculates the proportion of plans that satisfy all commonsense or hard constraints among the total number of tested plans.

In this study, we utilized three base prompting strategies for the Travel Planning task: Direct Prompting, Chain-of-Thought (CoT), and Step-Back Reasoning. 
We incorporated enhancements to the reference information, as recommended in \cite{singh2024personal}, to improve the effectiveness of the prompts.
Specific structural components for various base strategies are mentioned in Tables~\ref{app:ex:travel_plan_direct_prompt} to \ref{app:ex:travel_plan_step_back_prompt}.
Table~\ref{app:ex:travel_plan_llm_judge_prompt} shows the instructions used for the LLM-as-a-Judge and Table~\ref{app:ex:travel_plan_sff_prompt} depicts the same for strategy fusion.
Table~\ref{app:ex:travel_plan_base_prompt_structure} and \ref{app:ex:travel_plan_judgement_prompt_structure} shows the final prompt construction using components/instructions mentioned earlier.
 

\subsection{TravelPlanner: SMaRT Reasoning Trace}\label{app:reasoning}

\begin{figure}[t]
\centering
  \includegraphics[scale=0.29]{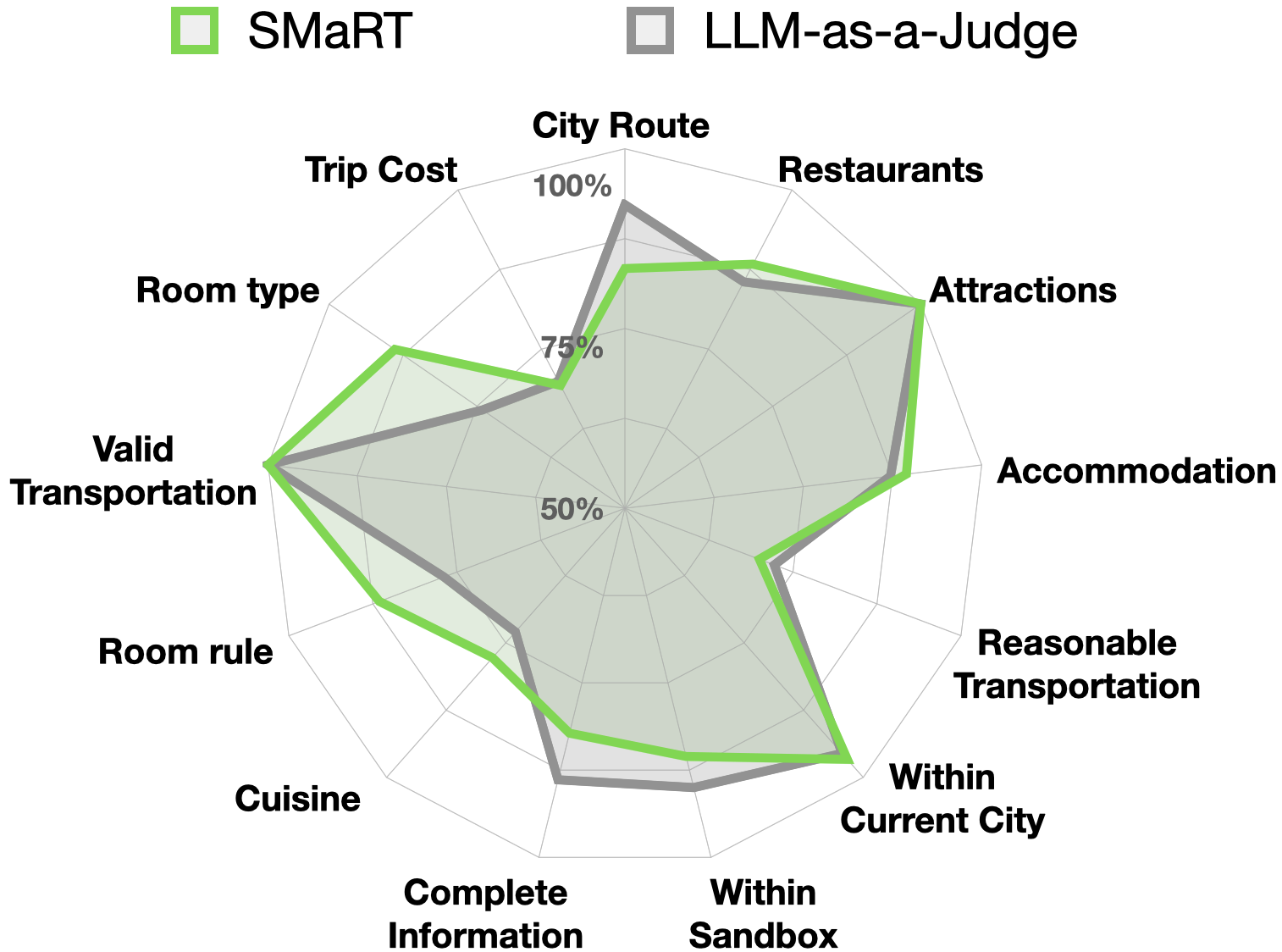}
  \caption {Radar plots illustrate the performance of SMaRT and LLM-as-a-Judge across various dimensions of the travel planner. SMaRT demonstrates better overall performance. }
    \label{fig:tp_base_strat_sff_comparison}
\end{figure}

As mentioned in the main text, the SMaRT also explains the reasoning behind specific actions taken during the fusion stage. For example, in the case of the travel planner, it clarifies why a particular component of a plan is selected from a base strategy, or why a new one is created. 
%
Each daily plan contains seven distinct components: current city, transportation, breakfast, lunch, dinner, attractions, and accommodation. 
On the validation split of the dataset (180 episodes), the Strategy Fusion process using the Gemini-1.5 model successfully created 56 plans, achieving a final pass rate of 31.11\%. 
These plans are distributed across varying durations, with 28 plans spanning three days, 16 plans spanning five days, and 12 plans spanning seven days.

Each successful plan integrates various sub-plan components, either selected from base strategies or newly generated using reference information provided during the fusion process. 
%
A significant portion of the final plan components (46.54\%) were common across all three base strategy plans, while others were included in two or just one of the base strategy outputs.
Additionally, 12.5\% of the successful plans featured entirely new components introduced during the strategy fusion stage.

\begin{table}[h]
\caption{Prompt for Direct strategy in TravelPlanner}\label{app:ex:travel_plan_direct_prompt}
\begin{analysisbox}

\textbf{BASIC\_TASK\_INSTRUCTIONS:}\\
You are a proficient travel planner.\\
You are provided with a Travel Query, Reference Information and illustration of Travel Plan.\\
Using the provided Reference Information and Travel Query, please give me a detailed Travel Plan.\\
Make sure to include specific information for each day of trip, such as\\
    * Flights/Self-Driving/Taxi: flight numbers (e.g., F0123456) with arrival and departure times or self-driving/taxi details. Do not combine 'self-driving' and 'flight' in the same trip\\
    * Restaurant: Suggest unique restaurants for Breakfast, lunch and dinner (e.g. restaurants\_XXXX)\\
    * Attractions: In the city of visit (e.g. attractions\_XXXX)\\
    * Accommodation (e.g. accommodations\_XXX) names for each day of the trip\\
Each day plan should include 'day', 'current\_city', 'transportation', 'breakfast', 'attraction', 'lunch', 'dinner', and 'accommodation'.\\
Strictly follow the format provided in the illustration plan.\\
The information for each plan should be derived only from the reference information.\\
Use the symbol '-' to indicates that information is unavailable/unnecessary.
Most importantly, ensure that the total trip cost, stays within the specified budget. \\
The travel plan should begin and end at the same city forming a closed circle.\\

\end{analysisbox}
\end{table}
\begin{table}[h]
\caption{Prompt for CoT strategy in TravelPlanner}\label{app:ex:travel_plan_cot_prompt}
\begin{analysisbox}

\textbf{STEP\_BY\_STEP\_THINKING\_FOR\_TASK:}\\
Thinking step-by-step to create a final travel plan:\\
    * Step 1: Understand the Query: create plan from Ithaca to Charlotte for 3 days, between March 8th and March 10th, 2022, for 7 people within \$30,200.\\
    * Step 2: Review Available Data: Access transportation, restaurants, attractions and accommodation options between Ithaca to Charlotte. Only use data from reference information.\\
    * Step 3: Create a Plan Outline: For each day of the trip, I need to fill in:\\
            - **Current City:** Indicating the starting and ending locations for that day.\\
            - **Transportation:** Flight, self-driving, or taxi details, where applicable.\\
            - **Breakfast:** Restaurant name.\\
            - **Attraction:** Choose one or more attractions per day.\\
            - **Lunch:** Restaurant name.\\
            - **Dinner:** Restaurant name.\\
            - **Accommodation:** Accommodation details for the night.\\
    * Step 4: Ensure Budget and constraint requirements: Ensure the total (for 7 people) stays within the budget (\$30,200) while ensuring the accommodation room rules, such as meeting the minimum consecutive nights' stay and adhering to the room type and rules specified in the travel query\\
    * Step 5: Build the Plan: Here is the travel plan in the required format using only the provided reference data with no repetition in restaurants and attractions\\

\end{analysisbox}
\end{table}
\begin{table}[h]
\caption{Prompts for Step Back strategy in TravelPlanner}\label{app:ex:travel_plan_step_back_prompt}
\begin{analysisbox}

\textbf{RESTAURANTS\_SHORTLISTING\_PROMPT:}\\
You are a proficient travel planner.\\
You are given a Travel Query along with a list of Restaurants Information. \\
Filter the restaurants that meet the travel criteria, ensuring no duplicates. \\
For each city in the itinerary, provide diverse selection of restaurants. \\
Do not create a travel plan, but only suggest restaurants.\\

\textbf{ACCOMMODATIONS\_SHORTLISTING\_PROMPT:}\\ 
You are a proficient travel planner.\\
You are given a Travel Query along with a list of Accommodation Information.\\
Filter the accommodations that meet the travel criteria, ensuring no duplicates. \\
For each city in the itinerary, provide a diverse selection of accommodations. \\
Do not create a travel plan, but only suggest accommodations.\\

\textbf{ATTRACTIONS\_SHORTLISTING\_PROMPT:}\\ 
You are a proficient travel planner.\\
You are given a Travel Query along with a list of Attractions Information. \\
Filter the attractions that meet the travel criteria, ensuring no duplicates. \\
For each city in the itinerary, provide a diverse selection of attractions.\\
Do not create a travel plan, but only suggest attractions.\\
\end{analysisbox}
\end{table}
\begin{table}[h]
\caption{Prompts for LLM-as-a-Judge strategy in TravelPlanner}\label{app:ex:travel_plan_llm_judge_prompt}
\begin{analysisbox}

\textbf{LLM-AS-A-JUDGE\_PROMPT:}\\
You are a proficient travel planner.\\
Identify the correct plan for a given Travel Query. \\
Provided a Travel Query and three different plans - (A), (B) and (C), you need to select the best plan which fulfills most of the constraints for Travel query. \\
You have to select the best plan by providing the index of the plan. \\
For example:\\
    * If the first plan (Plan (A)) is the best, respond with index (A)\\
    * If the second plan (Plan (B)) is the best, respond with index (B)\\
    * If the third plan (Plan (C)) is the best, respond with index (C) \\

\textbf{CONSTRAINTS\_TO\_FOLLOW:}\\
\#\# Constraints for Travel Query:\\
Environment Constraints:\\
- Unavailable Transportation: Verify if all transportation modes between cities are available.\\
- Attractions: No need to verify the availability or prices of the attractions.\\
\\
Commonsense Constraints:\\
- Budget: Total trip cost is within the specified budget.\\
- Complete Information: Ensure no key information is missing (e.g., accommodations, activities, transportation [to and fro], restaurants [breakfast, lunch, dinner]).\\
- Diverse Restaurants and Attractions: Avoid repetition of restaurant and attraction choices.\\
- Non-conflicting Transportation: Ensure consistent and logical transportation choices within the trip, avoiding conflicting options like combining 'self-driving' and 'flight' in the same trip. The travel plan should begin and end at the same city forming a closed circle.\\
- Minimum Nights Stay: Verify that required minimum stay durations are respected.\\
\\
Hard Constraints (only if present in the travel query):\\
- Budget: Total trip cost is within the specified budget.\\
- Room Rules: Check adherence to room rules (e.g., no parties, smoking, children under 10, pets, or visitors).\\
- Room Type: Ensure that specified room types (e.g., Entire Room, Private Room) are met.\\
- Cuisine: Verify that the plan includes the desired cuisines (e.g., Chinese, Italian, etc.).\\
- Transportation: Comply with any transportation restrictions.\\

\end{analysisbox}
\end{table}
\begin{table}[h]
\caption{Prompts for Strategy Fusion in TravelPlanner}\label{app:ex:travel_plan_sff_prompt}
\begin{analysisbox}

\textbf{STRATEGY\_FUSION\_INSTRUCTIONS:}\\
You are a proficient planner assistant that can create a correct plan for a given travel query. \\
\textbf{Given 3 version of planning, please provide a final plan which best helps in fulfilling all the constraints for Travel query.}\\
\textbf{You can either select best plan or cross and mix various components from the plans shared or use reference information to create final travel plan.} \\
You must create a plan which follows the same format as provided plans.
The provided plan should adhere to the following:\\
    - Daily Breakdown: Each day’s plan should include the following fields:\\
        * Day: Specify the day of the trip.\\
        * City: Current city of the traveler.\\
        * Transportation: Include details such as flight numbers and times, or self-driving/taxi information. Ensure no conflicts (e.g., don’t mix flights and self-driving in the same trip).\\
        * Meals: Provide unique suggestions for breakfast, lunch, and dinner (different restaurants each day).\\
        * Attractions: Suggest one or more attractions in the city for that day.\\
        * Accommodation: Name the accommodation, respecting any minimum night stay or house rules.\\
    Ensure the trip stays within budget and forms a closed circle, starting and ending in the same city. Use the symbol '-' if any information is unavailable or unnecessary.\\

\end{analysisbox}
\end{table}
\begin{table}[h]
\caption{Prompt Structure for various base strategies in TravelPlanner}
\label{app:ex:travel_plan_base_prompt_structure}
\begin{analysisbox}

\textbf{DIRECT\_PROMPT\_FOR\_TRAVEL\_PLANNING}:
\{BASIC\_TASK\_INSTRUCTIONS\}\\
\#\# Travel Query\\
\{travel\_query\}\\
\#\# Reference information \\
\{reference\_information\}\\
\#\# Illustration Travel Plan \\
\{ILLUSTRATION\_TRAVEL\_QUERY\}\\
\{ILLUSTRATION\_TRAVEL\_PLAN\}\\
\#\# Illustration ends \\

\textbf{COT\_PROMPT\_FOR\_TRAVEL\_PLANNING}:
\{BASIC\_TASK\_INSTRUCTIONS\}\\
Let’s think step-by-step to guide the planner for creating a travel plan.\\
The planner should consider all necessary details systematically, ensuring that all details are logically sequenced and aligned with the requested constraints in the travel query.  \\
\#\# Travel Query\\
\{travel\_query\}\\
\#\# Reference information \\
\{reference\_information\}\\
\#\# Illustration Travel Plan \\
\{ILLUSTRATION\_TRAVEL\_QUERY\}\\
\{STEP\_BY\_STEP\_THINKING\_FOR\_TASK\}\\
\{ILLUSTRATION\_TRAVEL\_PLAN\}\\
\#\# Illustration ends \\

\textbf{STEP\_BACK\_PROMPT\_FOR\_TRAVEL\_PLANNING}:
\{BASIC\_TASK\_INSTRUCTIONS\}\\
\#\# Travel Query\\
\{travel\_query\}\\
\#\# Reference information \\
\{reference\_information\} \# Obtained by shortlisting through SHORTLISTING\_PROMPT \\
\#\# Illustration Travel Plan \\
\{ILLUSTRATION\_TRAVEL\_QUERY\}\\
\{STEP\_BY\_STEP\_THINKING\_FOR\_TASK\}\\
\{ILLUSTRATION\_TRAVEL\_PLAN\}\\
\#\# Illustration ends \\

\end{analysisbox}
\end{table}
\begin{table}[h]
\caption{Prompt Structure for judgement-based strategies in TravelPlanner}\label{app:ex:travel_plan_judgement_prompt_structure}
\begin{analysisbox}
\textbf{LLM-AS-A-JUDGE\_PROMPT\_FOR\_TRAVEL\_PLANNING}:\\
\{LLM-AS-A-JUDGE\_PROMPT\}\\
\#\# Travel Query\\
\{travel\_query\}\\
\\
\{CONSTRAINTS\_TO\_FOLLOW\}\\
\\
\#\# Plans \\
Plan (A): \\
\{plan\_A\}\\
\\
Plan (B): \\
\{plan\_B\}\\
\\
Plan (C): \\
\{plan\_C\}\\
\\
Do not create a travel plan, but ONLY suggest the index of best plan without any reasoning.\\

\textbf{STRATEGY\_FUSION\_PROMPT\_FOR\_TRAVEL\_PLANNING}:\\
\{STRATEGY\_FUSION\_INSTRUCTIONS:\}\\
\#\# Travel Query\\
\{travel\_query\}\\
\\
\{CONSTRAINTS\_TO\_FOLLOW\}\\
\\
\#\# Plans \\
Plan (A): \\
\{plan\_A\}\\
\\
Plan (B): \\
\{plan\_B\}\\
\\
Plan (C): \\
\{plan\_C\}\\
\\
\#\# Reference information \\
\{reference\_information\}

\end{analysisbox}
\end{table}
\begin{table}[h]
\caption{Travel Planner: SMaRT Reasoning Trace}\label{app:tab:reasoning}
\begin{analysisbox}

\textbf{Explanation:}
\\
\\
This plan combines the best elements of Plans A, B, and C while adhering to all constraints:

\textbf{Transportation:} Uses the flights from Plans A and C, ensuring a closed circle trip with same mode of transport.

\textbf{Accommodation:} Uses the accommodations from Plans A and B, respecting minimum night stays and house rules.

\textbf{Attractions:}  Combines attractions from Plans B and C, ensuring diversity and avoiding repetition.

\textbf{Meals:}  Uses a mix of restaurants from Plans A, B, and C, offering diverse cuisines and avoiding repetition.

\textbf{Budget:}  The total cost of the trip is estimated to be around \$1,300, well within the \$1,500 budget.
\\
\\
\textbf{Breakdown of Costs:}
\\\\
\textbf{Flights:} \$160 (St. Louis to Washington) + \$146 (Washington to St. Louis) = \$306

\textbf{Accommodation:} \$209 (accommodations 1224) + \$411 (accommodations 3937) = \$620

\textbf{Meals:}  Estimated at \$200 (based on average costs from the reference information).

\textbf{Attractions:}  Estimated at \$50 (based on average attraction costs).
\\\\
\textbf{Total Estimated Cost:} \$306 + \$620 + \$200 + \$50 = \$1,176
\\
This plan provides a comprehensive and budget-friendly itinerary for a 3-day trip to Washington, starting and ending in St. Louis. 
\end{analysisbox}
\end{table}


\section{Further Experiments on Various Model Sizes for Base Strategies and SMaRT}\label{app:mix_models}

To further evaluate the effectiveness of the proposed framework, we conducted ablation studies using a smaller, open-source model: Gemma-2-9B \cite{gemma22024}(including their ``instruct'' variant). 
Among the datasets considered, the TravelPlanner multi-day itinerary planning task emerged as the most intricate due to its inherent complexity, making it an ideal candidate for assessing the robustness of our strategies.
As detailed in Table \ref{tab:travelplanner_val_split_results_gemma_9b}, we evaluated various strategies on the validation split of the TravelPlanner dataset using our custom prompts. 
Direct few-shot prompting achieved a best baseline final pass rate of 9.44\%.
By leveraging the LLM-as-a-Judge paradigm, we observed a performance improvement, elevating the pass rate to 10.56\%.
Notably, the Strategy Fusion delivered a significant performance enhancement, achieving an additional gain of 2.22\%, culminating in an overall pass rate of 12.78\%. 
This demonstrates the efficacy of SMaRT in synthesizing diverse strategies, even with resource-constrained models, thereby validating its generalizability and adaptability across different computational scales.

\begin{table}[h]
\caption{Performance indicators LLM agent using strategies on TravelPlanner's validation split with Gemma-2-9B model. Best results are shown in \textbf{bold}.}\label{tab:travelplanner_val_split_results_gemma_9b}
\centering
\resizebox{0.7\columnwidth}{!}{
\begin{tabular}{l || cccccc}
\hline

\multirow{3}{*}{\textbf{Strategy}} & \multicolumn{1}{c}{\multirow{3}{*}{
    \begin{tabular}[c]{@{}c@{}}\textbf{Delivery} \\ \textbf{Rate}\end{tabular}}} & \multicolumn{2}{c}{\textbf{Commonsense}} & \multicolumn{2}{c}{\textbf{Hard Constraint}} & \multicolumn{1}{c}{\multirow{3}{*}{\begin{tabular}[c]{@{}c@{}}\textbf{Final}\\ \textbf{Pass Rate}\end{tabular}}} \\ 
\multirow{3}{*}{} & \multicolumn{1}{c}{} & \multicolumn{2}{c}{\textbf{Pass Rate}} & \multicolumn{2}{c}{\textbf{Pass Rate}}  \\ \cline{3-6} 
\multirow{3}{*}{} & \multicolumn{1}{c}{} & \multicolumn{1}{c}{\textbf{Micro}} & \multicolumn{1}{c}{\textbf{Macro}} & \multicolumn{1}{c}{\textbf{Micro}} & \multicolumn{1}{c}{\textbf{Macro}} \\ \hline \hline

\textbf{Direct}& \textbf{100}	& 74.06	& 22.78	& 25.33	& 12.78	& 9.44\\
\textbf{CoT}&\textbf{100}	& \textbf{79.10}	& 13.88	& \textbf{47.06}	& \textbf{27.78}	& 5.56\\
\textbf{Step-Back}& \textbf{100}	& 72.50	& 15.56	& 26.32	& 18.89	& 7.78 \\ \hline
\textbf{LLM-as-a-Judge}& \textbf{100}	& 78.89	& \textbf{27.78}	& 40.88	& 26.67	& 10.56\\
\textbf{SMaRT}$_{\text{ours}}$ & \textbf{100}	& 76.74	& \textbf{27.78}	& 39.42	& 21.67	& \textbf{12.78}\\ \hline \bottomrule

\end{tabular}
}
\end{table}

Although Strategy Fusion demonstrates the ability to produce superior plans even with smaller models such as Gemma-2-9B, its performance metrics remain significantly lower than those achieved with larger, state-of-the-art models, as evident in Table \ref{tab:travelplanner_val_split_results_gemma_9b}.
To bridge this gap, we explored a hybrid approach, leveraging the complementary strengths of efficient small-scale open source models and stronger state-of-the-art models. 
Specifically, base strategy plans were generated using Gemma-2-9B, while the fusion phase was performed by the Gemini-1.5 model. 
This setup highlighted the adaptability of our framework compared to the conventional LLM-as-a-Judge paradigm, which merely selects from pre-generated plans without improving them.

Table \ref{tab:travelplanner_val_split_results_gemma_gemini} details the performance metrics across all strategies. 
While the LLM-as-a-Judge setup with Gemini-1.5 achieved a final pass rate of 10.56\%, identical to that obtained with Gemma-2-9B as the judge of its own base plans, SMaRT significantly outperformed this baseline. 
By integrating information from base plans and addressing missing or incorrect elements, SMaRT achieved a remarkable final pass rate of 22.78\%, demonstrating a nearly 2x performance improvement over the best LLM-as-a-Judge configuration.
This result underscores the transformative potential of SMaRT in combining the efficiencies of smaller models with the advanced capabilities of larger models, paving the way for resource-optimized, high-performing frameworks.

\begin{table}[h]
\caption{Performance indicators LLM agent using strategies on TravelPlanner's validation split with combination of models. Best results are shown in \textbf{bold}.}\label{tab:travelplanner_val_split_results_gemma_gemini}
\centering
\resizebox{0.7\columnwidth}{!}{
\begin{tabular}{l || cccccc}
\hline
\multirow{3}{*}{\textbf{Strategy}} & \multicolumn{1}{c}{\multirow{3}{*}{
    \begin{tabular}[c]{@{}c@{}}\textbf{Delivery} \\ \textbf{Rate}\end{tabular}}} & \multicolumn{2}{c}{\textbf{Commonsense}} & \multicolumn{2}{c}{\textbf{Hard Constraint}} & \multicolumn{1}{c}{\multirow{3}{*}{\begin{tabular}[c]{@{}c@{}}\textbf{Final}\\ \textbf{Pass Rate}\end{tabular}}} \\ 
\multirow{3}{*}{} & \multicolumn{1}{c}{} & \multicolumn{2}{c}{\textbf{Pass Rate}} & \multicolumn{2}{c}{\textbf{Pass Rate}}  \\ \cline{3-6} 
\multirow{3}{*}{} & \multicolumn{1}{c}{} & \multicolumn{1}{c}{\textbf{Micro}} & \multicolumn{1}{c}{\textbf{Macro}} & \multicolumn{1}{c}{\textbf{Micro}} & \multicolumn{1}{c}{\textbf{Macro}} \\ \hline \hline

\textbf{Direct}$^{Gemma-2-9B}$ & \textbf{100} & 74.06	& 22.78	& 25.33	& 12.78	& 9.44\\
\textbf{CoT}$^{Gemma-2-9B}$&\textbf{100}	& 79.10	& 13.88	& 47.06	& 27.78	& 5.56\\
\textbf{Step-Back}$^{Gemma-2-9B}$& \textbf{100}	& 72.50	& 15.56	& 26.32	& 18.89	& 7.78 \\ \hline
\textbf{LLM-as-a-Judge}$^{Gemini-1.5}$& \textbf{100}	& 78.54	& 23.89	& 42.33	& 23.33	& 10.56\\
\textbf{SMaRT}$_{\text{ours}}^{Gemini-1.5}$ & \textbf{100}	& \textbf{83.61}	& \textbf{42.78}	& \textbf{49.76}	& \textbf{38.33}	& \textbf{22.78}\\ \hline \bottomrule

\end{tabular}
}
\end{table}


\end{document}